\documentclass[10pt,twocolumn,letterpaper]{article}

\usepackage{cvpr}
\usepackage{times}
\usepackage{epsfig}
\usepackage{graphicx}
\usepackage{amsmath}
\usepackage{amssymb}
\usepackage{subfigure}

\newcommand{\tabincell}[2]{\begin{tabular}{@{}#1@{}}#2\end{tabular}}
\usepackage{multirow}
\usepackage[hyphens]{url}
\usepackage{amsthm}
\usepackage{enumitem}

\newtheorem{thm}{Theorem}
\newtheorem{assumption}{Assumption}

\hypersetup{pagebackref=true,breaklinks=true,letterpaper=true,colorlinks,bookmarks=false}

\usepackage{listings}
\usepackage{color}

\definecolor{dkgreen}{rgb}{0,0.6,0}
\definecolor{gray}{rgb}{0.5,0.5,0.5}
\definecolor{mauve}{rgb}{0.58,0,0.82}

\cvprfinalcopy

\def\cvprPaperID{4457} 

\ifcvprfinal\fi
\begin{document}

\title{Towards Unified INT8 Training for Convolutional Neural Network}

\author{Feng Zhu$^{1}$ \quad
Ruihao Gong$^{1, 2}$\quad
Fengwei Yu$^{1}$ \quad
Xianglong Liu$^{2}$\thanks{indicates corresponding author} \\
Yanfei Wang$^{1}$ \quad
Zhelong Li$^{1}$ \quad
Xiuqi Yang$^1$ \quad
Junjie Yan$^1$\\
$^1$SenseTime Research\\
$^2$ State Key Laboratory of Software Development Environment, Beihang University\\
{\tt\small 
\{zhufeng1, yufengwei, wangyanfei, lizhelong, yangxiuqi, yanjunjie\}@sensetime.com } \\
{\tt\small 
	\{gongruihao, xlliu\}@nlsde.buaa.edu.cn}
}

\maketitle

\begin{abstract}
    Recently low-bit (e.g., 8-bit) network quantization has been extensively studied to accelerate the inference. Besides inference, low-bit training with quantized gradients can further bring more considerable acceleration, since the backward process is often computation-intensive. Unfortunately, the inappropriate quantization of backward propagation usually makes the training unstable and even crash. There lacks a successful unified low-bit training framework that can support diverse networks on various tasks. In this paper, we give an attempt to build a unified 8-bit (INT8) training framework for common convolutional neural networks from the aspects of both accuracy and speed. First, we empirically find the four distinctive characteristics of gradients, which provide us insightful clues for gradient quantization. Then, we theoretically give an in-depth analysis of the convergence bound and derive two principles for stable INT8 training. Finally, we propose two universal techniques, including Direction Sensitive Gradient Clipping that reduces the direction deviation of gradients and Deviation Counteractive Learning Rate Scaling that avoids illegal gradient update along the wrong direction. The experiments show that our unified solution promises accurate and efficient INT8 training for a variety of networks and tasks, including MobileNetV2, InceptionV3 and object detection that prior studies have never succeeded. Moreover, it enjoys a strong flexibility to run on off-the-shelf hardware, and reduces the training time by 22\% on Pascal GPU without too much optimization effort. We believe that this pioneering study will help lead the community towards a fully unified INT8 training for convolutional neural networks.
\end{abstract}

\section{Introduction}
Deep convolutional neural networks (DCNNs) have achieved remarkable success in many fields, such as computer vision, natural language processing, information retrieval, etc. However, training and deploying DCNNs usually require a large amount of computational cost and power consumption, which is greatly challenging the extensive applications in industry. As a result, many recent studies have been focusing on how to accelerate the inference of neural networks by fixed-point quantization on weights or activations~\cite{PACT,PostTraining4bit,GoogleCVPR2018,Whitepaper,LearnInterval,zhou2016dorefa,Apprentice,HAQ,INQ,TSQ,hou2018lossaware}, and design dedicated hardware utilizing the efficient integer arithmetic~\cite{EIE,cambricon,tpu,ascend310}. The successful progress surprisingly shows that the bit-width can be reduced to extremely low such as 4-bit while bringing quite little hurt to the accuracy for inference \cite{dsq,quantization_networks,lsq}.

\begin{figure}[t!]
\centering
\includegraphics[width=1\linewidth]{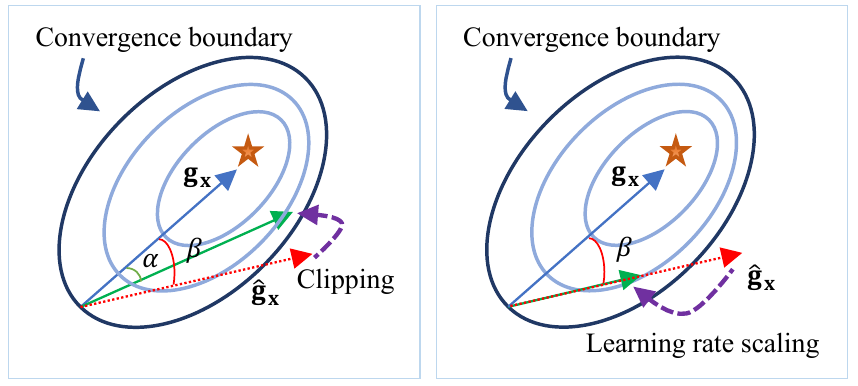}
\caption{The fundamental idea of our unified INT8 training. $\mathbf{g_x}$ and $\mathbf{\hat{g}_x}$ represent the original float gradient and the quantized one, respectively. $\alpha$ and $\beta$ represent different direction deviations that quantization brings. The red lines present crash cases when the direction deviation is large. The left subfigure indicates that clipping gradient properly to reduce direction deviation within the convergence boundary can avoid crash. The right subfigure points out that controlling learning rate (step size) could promise a stable parameter updating by counteracting negative effect of deviation. }
\vspace{-0.1in}
\label{fig:optimize}
\end{figure}
Besides inference, low-bit training can also promise considerable acceleration, which further quantizes gradients and utilizes low-bit efficient compute kernel for both the forward and backward propagation. As analyzed in \cite{cnn-benchmark}, the computation of backward propagation occupies more time than that of forward propagation. So accelerating the training utilizing low-bit quantization has greater potential when considering the backward process. There has existed 16-bit floating-point (FP16) training, which proves the feasibility of low-bit training \cite{MPT,DFP,Flexpoint}. But it is restricted to limited advanced GPUs based on Turing or Volta architecture. Compared with FP16, the 8-bit integer (INT8) operation is widely supported by general GPUs based on Turing, Volta and even low-end Pascal architectures. Besides, the 8-bit integer arithmetic is theoretically and practically 2$\times$ faster than FP16 and 4$\times$ faster than FP32. Therefore, INT8 training enjoys better efficiency, lower power consumption and better versatility on off-the-shelf hardware.

Despite the attractive benefits, when quantizing gradients to 8-bit, the normal training tends to become unstable, since the distortion of gradients easily misleads the direction of training and causes crash of optimization. This definitely makes INT8 training very difficult, especially for the deep networks. Currently only a few studies have attempted to solve this problem~\cite{zhou2016dorefa,wage,wageubn,banner2018scalable,fp8training,PerTensorFX}. Unfortunately, all of them just tested limited quantization-friendly networks with high redundancy, and usually require complex structure adjustment or introduce additional operation to reduce quantization error, while significantly increasing the computational complexity. Besides, most of these works lack the theoretical analysis on the ad-hoc tricks, and even worse, none of them reports the practical speedup in the real-world case. All these reasons make the existing INT8 training methods stay far away from the practicality without the universal design.

To build a robust and unified INT8 training framework, we conduct deeper explorations in the challenges of gradient quantization. We empirically find that the distribution of gradients owns four special characteristics: sharp and wide, evolutionary, depth-specific and structure-specific. These unique characteristics make gradient quantization quite different from the naive quantization on weights or activations, and INT8 training more difficult to be stabilized. It is important to understand the behaviors and effects of quantized gradient in the convergence of the training. Therefore, we theoretically establish the convergence bound with respect to the gradient quantization error and the learning rate.

Based on the special characteristics and the theoretical analysis, we propose two universal techniques: Direction Sensitive Gradient Clipping and Deviation Counteractive Learning Rate Scaling to stabilize the INT8 training. The Direction Sensitive Gradient Clipping minimizes the direction deviation by pursuing an appropriate clipping as the training process evolves. Sometimes even if the clipping helps reduce the quantization error, it may still suffer from the accumulated gradient deviations across deep layers. To eliminate this effect, the Deviation Counteractive Learning Rate Scaling is further devised to promise stable parameter updating. The fundamental idea of our method is shown in Figure \ref{fig:optimize}. Extensive experiments on a variety of network structures and tasks prove the superiority and versatility of our method.

Our contribution can be summarized as below:
\begin{itemize}
    \item We observe four special characteristics on the gradient distribution: sharp and wide, evolutionary, depth-specific and structure-specific, which cause the larger quantization error of gradients.
    \item We theoretically provide the convergence bound of INT8 training, and respectively devise two universal techniques that can stabilize the INT8 training.
    \item We are the first to achieve stable INT8 training of various networks such as MobileNetV2/InceptionV3 and various tasks such as object detection, with comparable accuracy to full-precision training.
    \item We build a flexible and unified INT8 training framework for various tasks using various networks, which can easily replace the original full-precision training.
    \item We are the first to complete practical acceleration of INT8 training on low-end GPUs with Pascal architecture, i.e., NVIDIA GeForce GTX 1080Ti, achieving about 22\% speedup without too much code optimization.
\end{itemize}

\section{Related Work}
Compared to huge amount of studies on accelerating inference by model quantization~\cite{XnorNet,LQNet,BinaryConnect,HAQ,RAD,DiscoveringLowPrecision}, there are few works exploring quantized training including backward propagation comprehensively. DoReFa-Net~\cite{zhou2016dorefa} quantizes gradients to 4 and 6 bits, but only experiments AlexNet with low precision gradient. WAGE~\cite{wage} and WAGEUBN~\cite{wageubn} quantize gradient to 8-bit integer, but they both incur considerable loss of accuracy (greater than $5\%$). RangeBN~\cite{banner2018scalable} and FP8 training \cite{fp8training} achieve accuracy comparable to full-precision models, but they both use floating-point number in gradients, which is not beneficial for hardware optimization to boost the speed.
Besides quantized training, most low-precision training research keeps gradient precision in 16-bit floating-point. Flexpoint~\cite{Flexpoint}, MPT~\cite{MPT} and DFP~\cite{DFP} all use 16-bit floating-point to train DNNs with accuracy comparable to full-precision model. To perform more efficient training of neural networks, INT8 training has more advantages over FP16 training.

\section{Unified INT8 Training}
In this paper, we aim to build a unified INT8 training framework, which utilizes 8-bit integer arithmetic to accelerate the expensive training process of deep neural networks including both the forward and backward propagation. 
\subsection{Preliminaries}
Symmetric uniform quantization is the most efficient scheme among existed quantization methods, due to its hardware-friendly computation. Therefore, to guarantee the acceleration performance, we build the INT8 training framework based on it. Given the data $x$ (i.e., weights, activations, and gradients) following in the range $(l, u)$ and a clipping value $c \in (0, \max(|l|, |u|)]$, the symmetric uniform quantization can be formulated as:
\begin{equation}
    \label{eq:quant}
    q = \mathtt{round}(\frac{\mathtt{clip}(x, c)}{s}),
\end{equation}
where $\mathtt{clip(x, c)=\min(\max(x, -c), c)}$, $s=\frac{c}{2^{8-1}-1}$ indicates the scaling factor to project the floating-point number to fixed-point 8-bit integer, and $q$ represents the quantized fixed-point number. Subsequently, the corresponding dequantized data $\hat{x}$ can be calculated by:
\begin{equation}
    \label{eq:dequant}
    \hat{x} = q \cdot s.
\end{equation}

Different from most prior studies that mainly focus on speeding up the inference (i.e., the forward propagation), our INT8 training framework attempts to further accelerate the backward propagation during the training stage, by applying quantization to the gradients. Namely, we pursue the quantize-dequantized gradients $\mathbf{\hat{g}}$ from full-precision gradients $\mathbf{g}$ in a proper way.

To ensure the quantized gradients maintain an unbiased expectation compared with the original ones, we adopt the stochastic rounding following \cite{pmlr-v37-gupta15}:
\begin{equation}
    \mathtt{round_s(x)} = \begin{cases}
    \lfloor x \rfloor, & \mathtt{w.p.} \quad 1-(x-\lfloor x \rfloor) \\
    \lfloor x \rfloor + 1, & \mathtt{w.p.} \quad x-\lfloor x \rfloor \\
    \end{cases}.
\end{equation}


Unfortunately, although the stochastic rounding technique limits the quantization error to some extent from the statistical view, the perturbation for each training iteration is still inevitable and harmful for convergence, whose reasons will be discussed in the following section.

\begin{figure}[tp!]
\centering

\subfigure[the accuracy curve]{
\includegraphics[width=.47\linewidth]{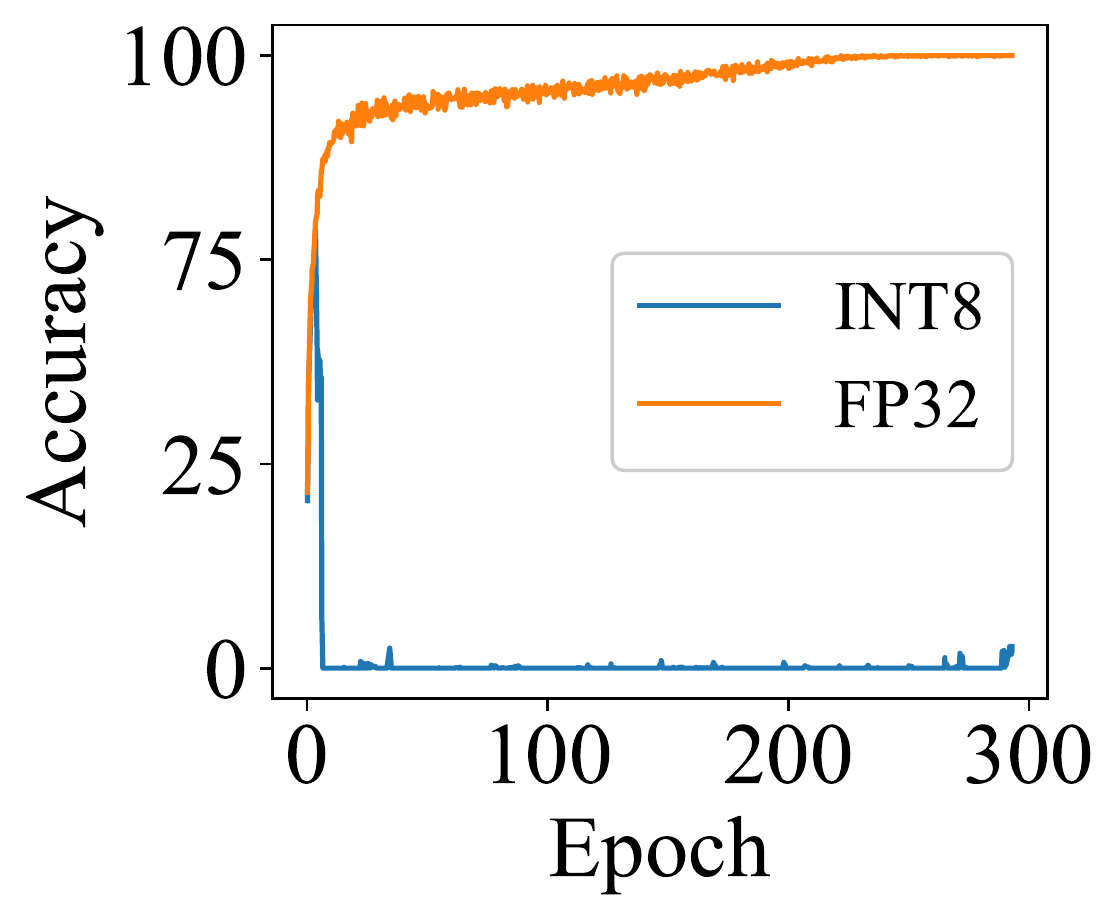}
}
\subfigure[the loss curve]{
\includegraphics[width=.47\linewidth]{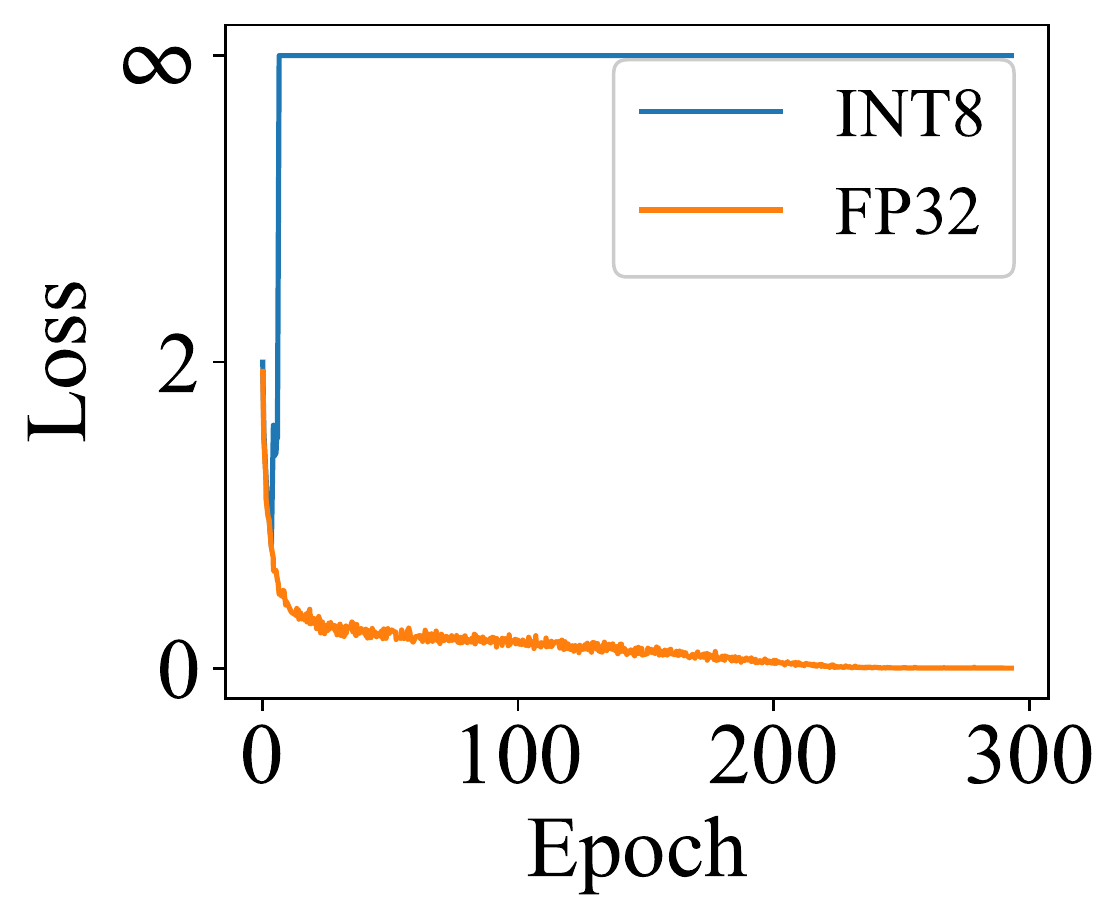}
}
\caption{Crashed training of MobileNetV2 on CIFAR-10 after quantizing gradients to 8-bit.}
\vspace{-0.1in}
\label{fig:crash}
\end{figure}

\subsection{Challenges of Gradient Quantization}
\label{section:challenges}
Gradients determine the direction of optimization and the magnitude of parameter update and thus play a critical role in pursuing high accurate models. In INT8 training, after we apply quantization to gradients, the perturbation introduces deviation to the optimization direction. Once the deviation accumulates to an unacceptable degree, the training process may be unstable and even crash, resulting in severe performance degradation. Figure \ref{fig:crash} shows our empirical observation that for some special network architectures like MobileNetV2, directly quantizing gradients causes a rapid crash of training.

\begin{figure}[tp!]
\subfigure[gradients are different from weights and activations]{
\includegraphics[width=1\linewidth]{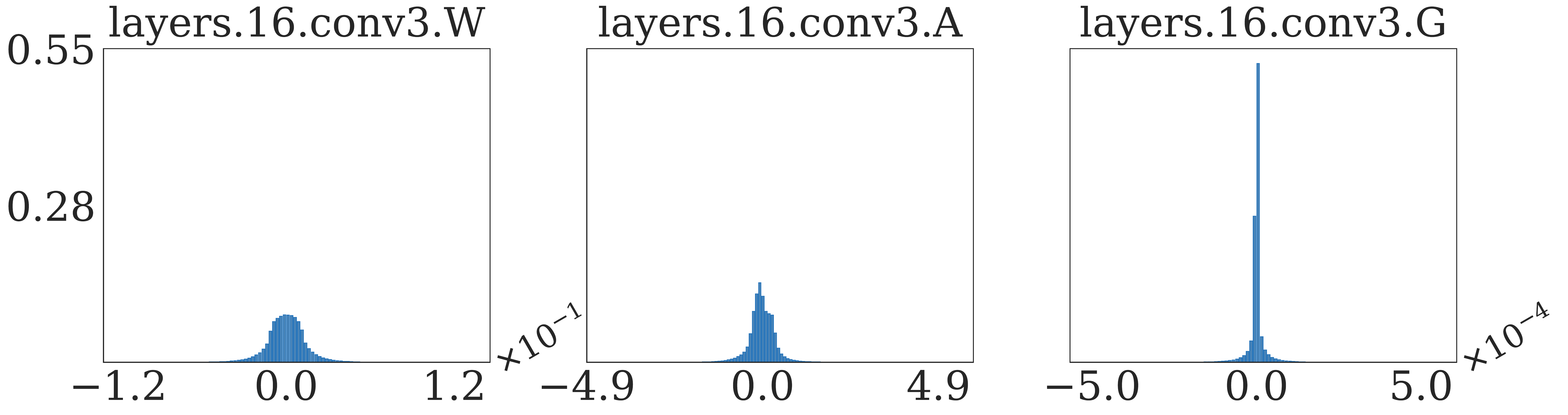}
}
\subfigure[gradients keep evolving during training]{
\includegraphics[width=1\linewidth]{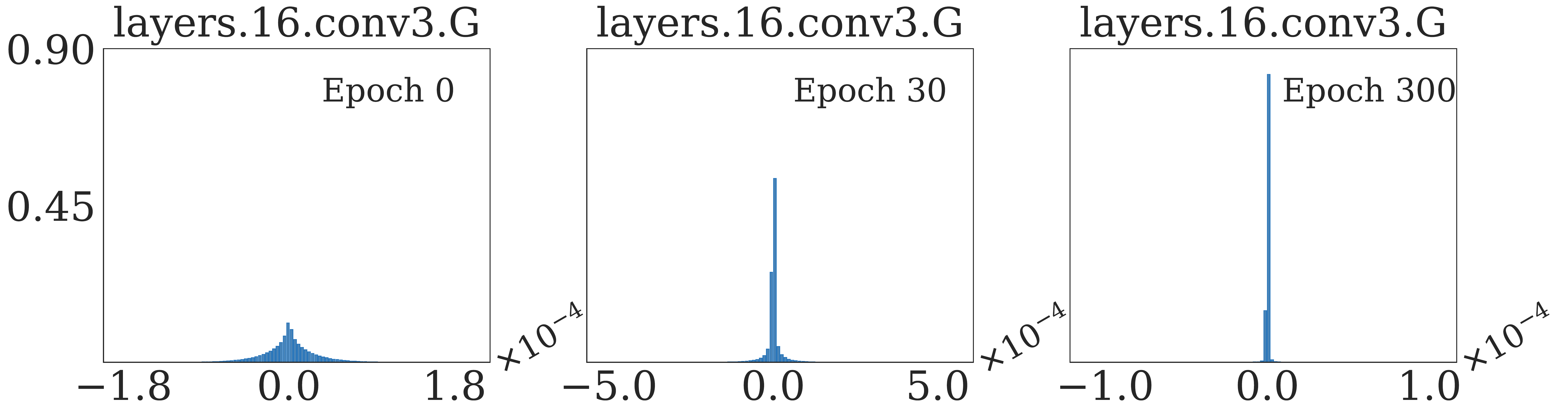}
}
\subfigure[gradients of different depths have have different patterns]{
\includegraphics[width=1\linewidth]{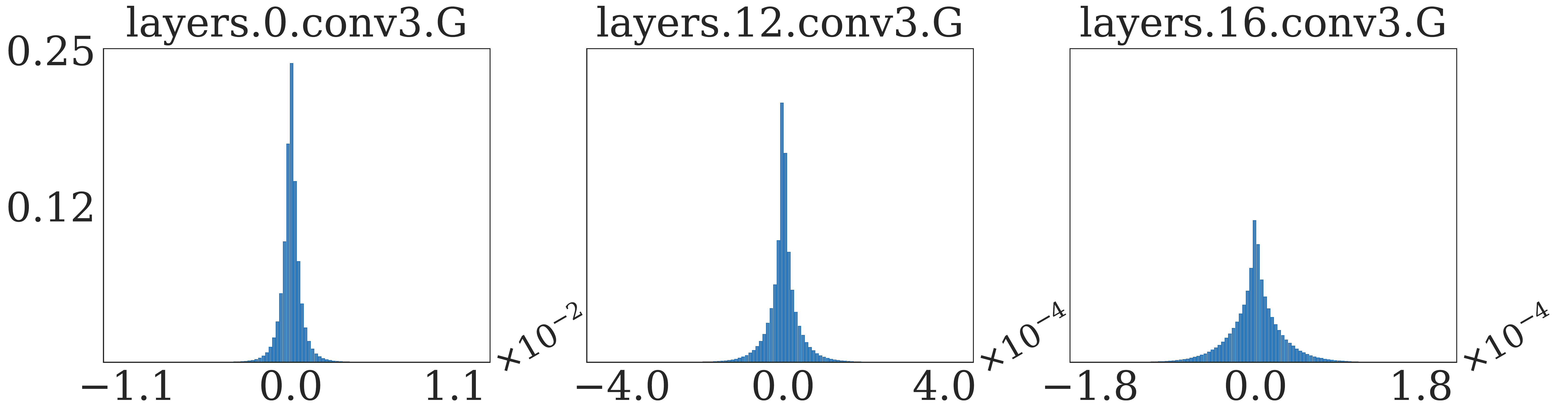}
}
\subfigure[gradients of different structures have different patterns]{
\includegraphics[width=\linewidth]{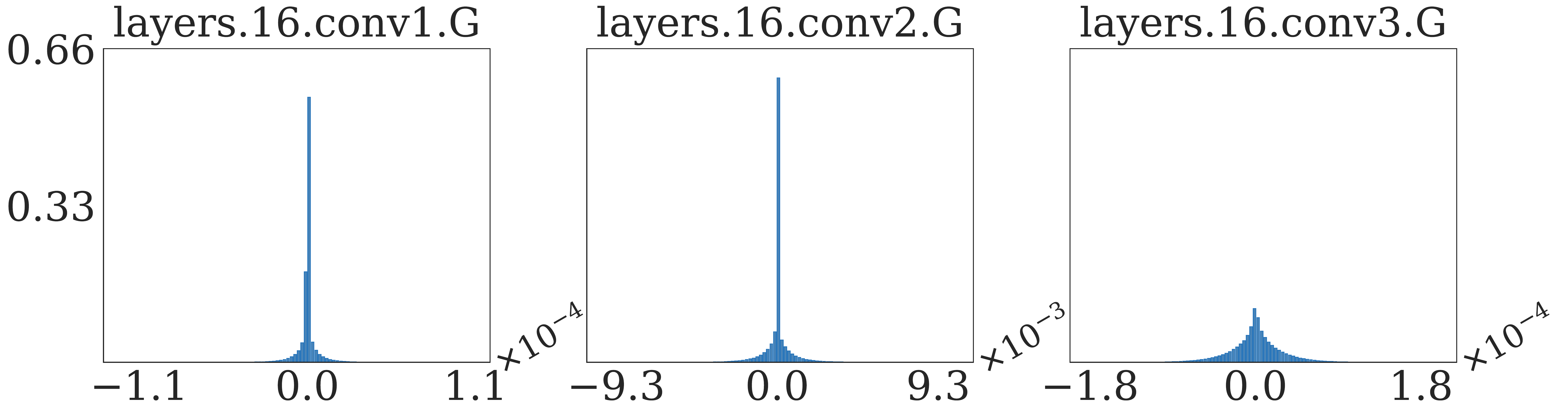}
}
\caption{Distributions of activations, weights and gradients with respect to different layers of MobileNetV2 and training iterations.}
\label{fig:data_distritbuion}
\vspace{-0.1in}
\end{figure}

To further investigate the essential reasons behind this phenomenon, we conduct detailed analysis on the distribution of gradients during training without gradient quantization, as shown in Figure \ref{fig:data_distritbuion}. We surprisingly observe that the gradients own the following unique characteristics:

\begin{itemize}
    \item[C1:] \textbf{Sharp and Wide.} As shown in Figure \ref{fig:data_distritbuion}(a), compared to weights and activations, gradients follow an unusual distribution that has more values concentrated around zero while a certain number of extreme values also exists. Therefore, the distribution curve is very sharp with small values taking the majority of gradients, but the range is relatively very wide. This makes many gradients quantized to zero and the quantization error significantly large when using uniform quantization.
    
    \item[C2:] \textbf{Evolutionary.} Figure \ref{fig:data_distritbuion}(b) depicts how the gradient distribution of the same layer evolves with respect to the training iterations. We can find that as the training goes on, the shape of gradient distribution becomes much sharper and narrower, which means it is impossible to fix the quantization settings throughout the training process, as we usually do for weights and activations, such as assuming the same clipping range in the whole training.
    
    \item[C3:]  \textbf{Depth-Specific.} Figure \ref{fig:data_distritbuion}(c) compares the  distribution of gradients in different layers. It is obvious that the distributions in the shallow layers are sharper with larger extreme values than the deeper layers. This means that the preceding layers of the deep neural networks often face more severe quantization loss.
    
    \item[C4:] \textbf{Structure-Specific.} As can be seen in Figure \ref{fig:data_distritbuion}(d), the gradients of layers with different structures present apparently different patterns. For MobileNetV2, the second convolutional layer in each block is of depth-wise structure. Its gradients own larger range and sharper shape even in the deeper block, making MobileNetV2 harder to quantize from the aspect of gradients.
\end{itemize}

Based on the above observations, we can conclude that the gradients differ from weights and activations largely, which inevitably causes an unstable training, when simply adopting the common quantization techniques for weights and activations. This means that we need certain techniques to take care of distinctiveness in gradient quantization, which brings great challenges to the real and unified INT8 training in practice.


Before turning to devise the desired techniques considering the speciality of gradients, we first attempt to understand the gradient's effect on the training stability, by theoretically revealing the connections between training convergence and gradient quantization. This will provide us a reliable clue to build the robust and unified INT8 training framework.


\subsection{Stabilize Training: A Theoretical Perspective}
\label{convergence_analysis}
As commonly used in the analysis of deep learning optimizers \cite{duchi2011adaptive,Adam,AMSGrad,Luo2019AdaBound}, the ability of convergence is usually evaluated by the regret $R(T)$. 
\begin{equation}
\label{eq:regret}
R(T) = \sum_{t=1}^{T} (f_t(\mathbf{w}_t)-f_t(\mathbf{w}^*)),
\end{equation}
where $T$ indicates the number of iterations. $\mathbf{w}_t \in \mathbb{S}$ is the parameter at time $t$ in the convex compact set $\mathbb{S}$, and $f_t(\mathbf{w}_t)$ denotes the corresponding loss function. The optimal parameter is represented by $\mathbf{w}^*$. If the average regret $\frac{R(T)}{T}$ approaches zero quickly as $T$ increases, the speed and ability of convergence can be guaranteed. 

Due to the complexity of the DCNNs, it is very difficult to directly analyze its behaviors. As the prior studies \cite{QSGD,DeeperUnderstanding,hou2018analysis,BCGD} do, we first make the following assumptions:
\begin{assumption}  \label{assumption:1} 
$f_t$ is convex;
\end{assumption}
\begin{assumption}  \label{assumption:2} 
$\forall \mathbf{w}_i, \mathbf{w}_j \in \mathbb{S}, \|\mathbf{w}_i-\mathbf{w}_j \|_\infty \leq D_\infty$.
\end{assumption}

Although the convexity assumption may not hold for deep networks, analysis based on this can provide reasonable and valuable insights for us, which has been proved in previous studies \cite{duchi2011adaptive,Luo2019AdaBound,hou2018analysis,BCGD}.

Taking the standard stochastic gradient descent algorithm into consideration, the optimization based on quantized gradient $\mathbf{\hat{g}}_t$ and learning rate $\eta_t$ can be formulated as:
\begin{equation}
    \mathbf{w}_{t+1} = \mathbf{w}_t - \eta_t\mathbf{\hat{g}}_t.
\end{equation}
Then we have the following theoretical finding (see the supplementary materials for detailed proof):
\begin{thm}
\label{thm:1}
If define the error of quantized gradients as $\mathbf{\epsilon}_t = \mathbf{g}_t-\mathbf{\hat{g}}_t$, then with assumption \ref{assumption:1} and \ref{assumption:2}, we have:
\begin{equation}
\label{eq:avg_regret}
    \frac{R(T)}{T} \leq \underbrace{\frac{d D_\infty^2}{2T\eta_{T}}\vphantom{\sum_{t=1}^{T}}}_{(1)} + \underbrace{ \frac{D_\infty}{T} \sum_{t=1}^{T} \| \mathbf{\epsilon}_t \|\vphantom{\sum_{t=1}^{T}}}_{(2)} + \underbrace{\frac{1}{T}\sum_{t=1}^{T} \frac{\eta_{t}}{2}\|\mathbf{\hat{g}}_{t}\|^2}_{(3)}.
\end{equation}
\end{thm}

We can find that the bound of average regret is dominated by three terms. Term (1) approaches zero as $T$ increases and thus can be ignored in gradient quantization. Term (2) indicates the quantization error of gradients greatly affects the ability to converge, and it is usually large, as analyzed in Section \ref{section:challenges}. For term (3), its magnitude is mainly influenced by the learning rate and l2-norm of quantized gradients. Based on the theoretical analysis, to stabilize INT8 training, we have two basic principles for designing better quantization techniques:  (1) reduce the quantization error of gradients; (2) scale down the learning rate. They are also very intuitive since, on the one hand, a lower quantization error means small deviation of optimization direction and thus avoids the training crash, on the other hand, it is a common sense that decreasing the learning rate gradually promises a better solution in the optimization.  


Now with the design principles, the question is how to devise the universal techniques for INT8 training, meanwhile take the characteristics of gradients into consideration. We respectively present two novel techniques: Direction Sensitive Gradient Clipping and Deviation Counteractive Learning Rate Scaling, which together lower the average regret bound and guarantee stable INT8 training.

\subsection{Direction Sensitive Gradient Clipping}
Considering the basic operation $\mathbf{z} = \mathbf{W}^\top \mathbf{a}$ in deep neural networks, the gradients of weights $\mathbf{g_W}$ actually can be calculated by $\mathbf{g_z}^\top \mathbf{a}$. From this aspect, the quantization error of $\mathbf{g_W}$ in \eqref{eq:avg_regret} mainly stems from that of activation gradients $\mathbf{g_z}$. Therefore, in our INT8 training we can mainly concern the quantization of $\mathbf{g_z}$, which will help control the error of quantized gradients in \eqref{eq:avg_regret}. For simplicity of notations, in the following discussion we directly use $\mathbf{g}$ to denote $\mathbf{g_z}$.


To minimize quantization error, previous works mainly seek the optimal clipping value $c$ in \eqref{eq:quant} by assuming certain data distribution, e.g. Gaussian distribution \cite{PostTraining4bit,HWGQ,truncated_gaussian,banner2018scalable,hou2018analysis,RAD}. However, according to the gradient characteristics C1 and C2 we discover, it is unpractical to make a common assumption for an evolutionary and unusual gradient distribution. To further prove this point, we do the Kolmogorov–Smirnov test with distribution parameter solved by maximum likelihood estimation, and report the KS-statistics that consistently reject the assumption that gradients obey any common distribution in Table \ref{table:hyp_test}.

\begin{table}[t!]
\caption{KS-statistics of gradient and weight with respect to different layers' conv3 in MobiletNetV2, the last column indicates the maximum value that can accept the hypothesis at significance level of 0.05.}
\label{table:hyp_test}
\centering
\small
\begin{tabular}{|c|c|c|c|c|c|}
\hline
\multicolumn{2}{|c|}{\multirow{2}{*}{Data}} & \multicolumn{3}{|c|}{Distribution} & \multirow{2}{*}{Critical value} \\
\cline{3-5}
\multicolumn{2}{|c|}{} & Gaussian  & Laplace & Student & \\
\hline
\hline
\multirow{2}{*}{layer0} & $\mathbf{g}$ & 0.1934 & 0.0790 & 0.2005 & 0.0012 \\
  & $\mathbf{w}$ & 0.0391 & 0.0721 & 0.1011 & 0.0765 \\
\hline
\multirow{2}{*}{layer8} & $\mathbf{g}$ & 0.2061 & 0.1091 & 0.2303 & 0.0024\\
 & $\mathbf{w}$ & 0.0294 & 0.0569 & 0.1084 & 0.0110 \\
\hline
\end{tabular}
\vspace{-0.14in}
\end{table}

To find the optimal clipping value $c$ without any assumption, a straightforward idea is to keep the quantized gradient consistent with the original one by gradient descent algorithm. Usually, one can model the consistency using the popular mean-square error (MSE). Unfortunately, due to characteristics C2 and C3 of gradients with huge discrepancy and fluctuation in their magnitudes, MSE makes the optimization vulnerable and unable to work under the same simple setting across various layers.

Therefore, to pursue the desired clipping values of different layers that promise stable training, we choose cosine distance to guide the learning of clipping values, which not only avoids the negative effect of the varied gradients' magnitudes, but also keeps the network optimization directions consistent:
\begin{equation}
\label{eq:cos_err}
    d_c = 1 -\cos(<\mathbf{g}, \mathbf{\hat{g}}>) = 1 - \frac{\mathbf{g} \cdot \mathbf{\hat{g}}}{|\mathbf{g}| \cdot |\mathbf{\hat{g}|}}
\end{equation}
where $\mathbf{g}$ and $\mathbf{\hat{g}}$ denote the original floating-point gradient and its quantize-dequantized counterpart. 

\begin{figure}[t!]
\centering
\subfigure[the accuracy curve]{
\includegraphics[width=.47\linewidth]{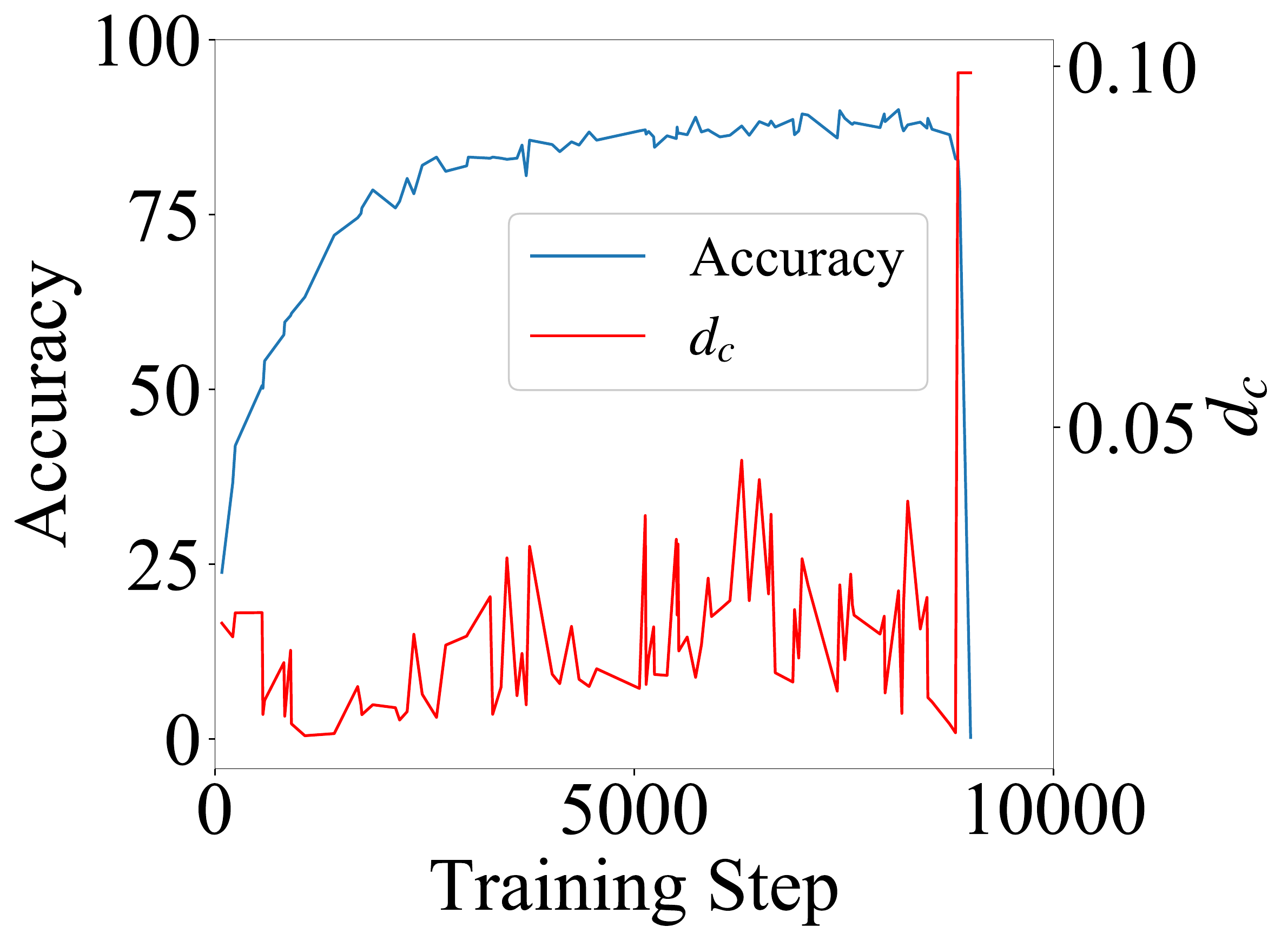}
}
\subfigure[the loss curve]{
\includegraphics[width=.47\linewidth]{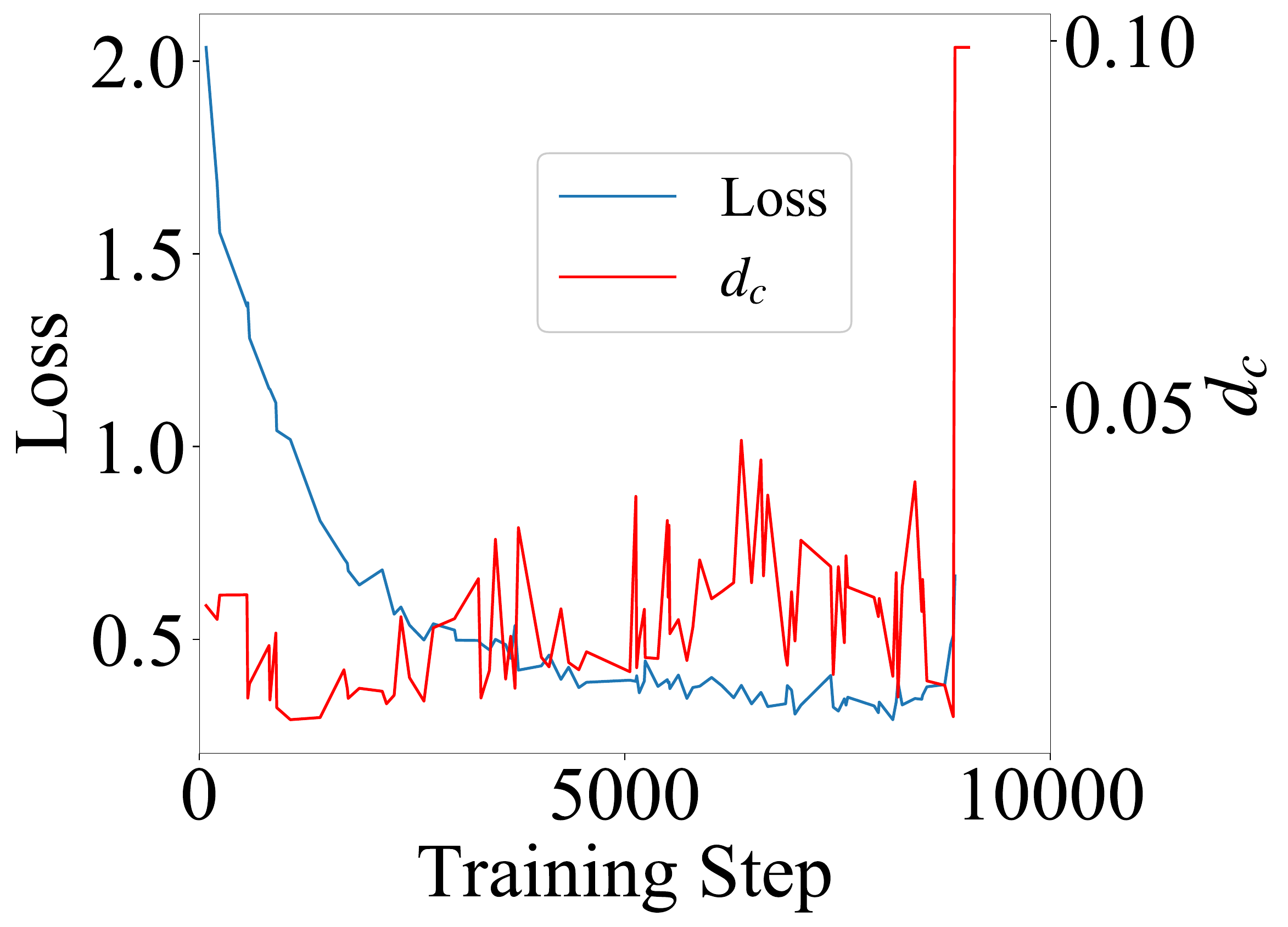}
}
\caption{Model crashes when $d_c$ exceeds limits.}\label{fig:CosAcc}
\vspace{-0.1in}
\end{figure}

The cosine distance measures the direction deviation of quantized gradients. As shown in Figure \ref{fig:CosAcc}, when $d_c$ increases to a certain level, the whole training crashes. There exists strong correlation between $d_c$ and training stability, which proves that cosine distance can effectively reflect the influence of gradient quantization on the convergence. By minimizing the deviation, we subsequently reduce term (2) in \eqref{eq:avg_regret}. Figure \ref{fig:effect}(a) shows the quantization error using different clipping values, where there exists an optimal clipping value that substantially reduces the cosine distance. 

\subsection{Deviation Counteractive Learning Rate Scaling}

\begin{figure}[t!]
\subfigure[effect of clipping]{
  \includegraphics[width=0.42\linewidth]{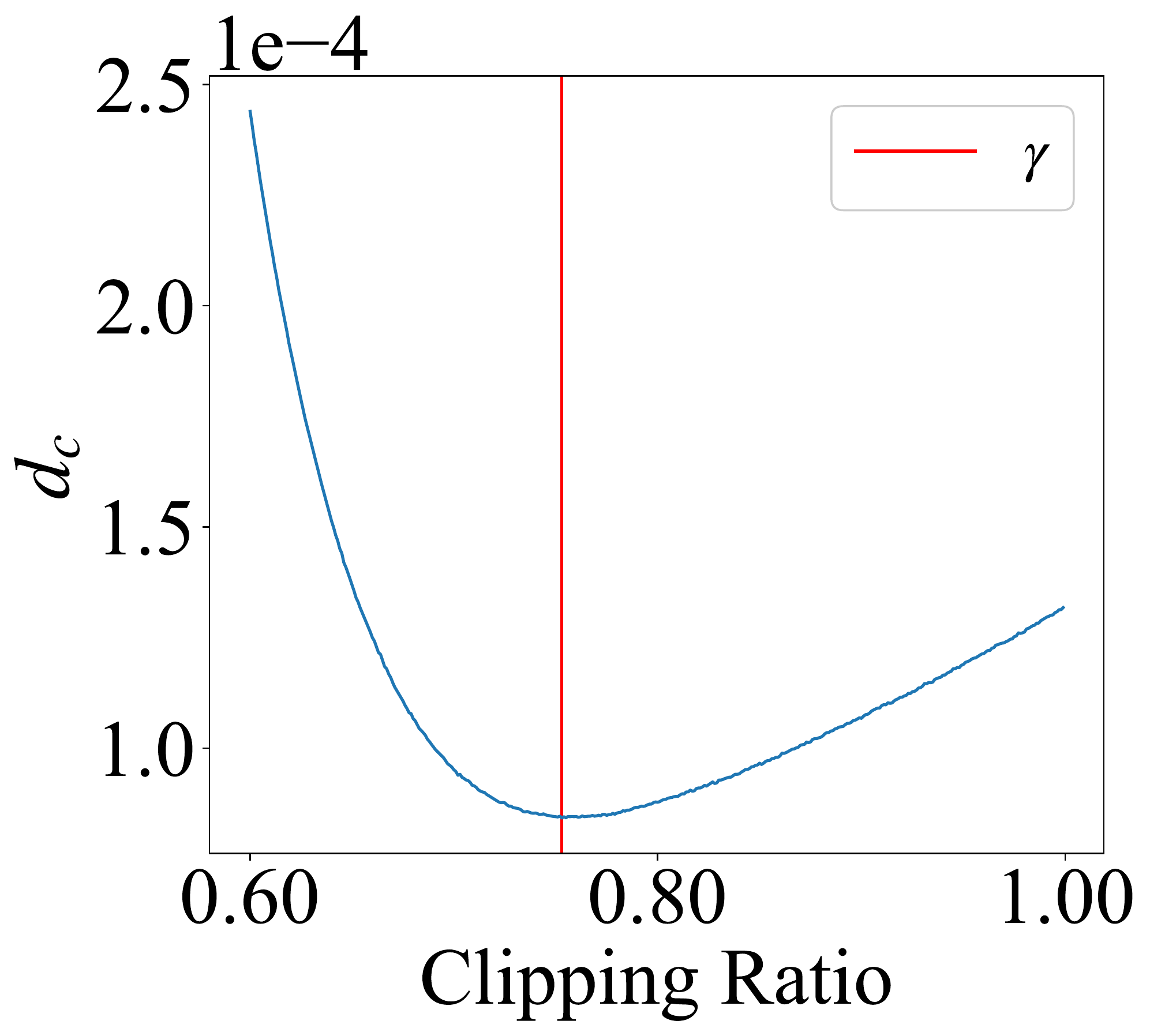}
}
\hspace{0.1in}
\subfigure[effect of scaling strategies]{
  \includegraphics[width=0.43\linewidth]{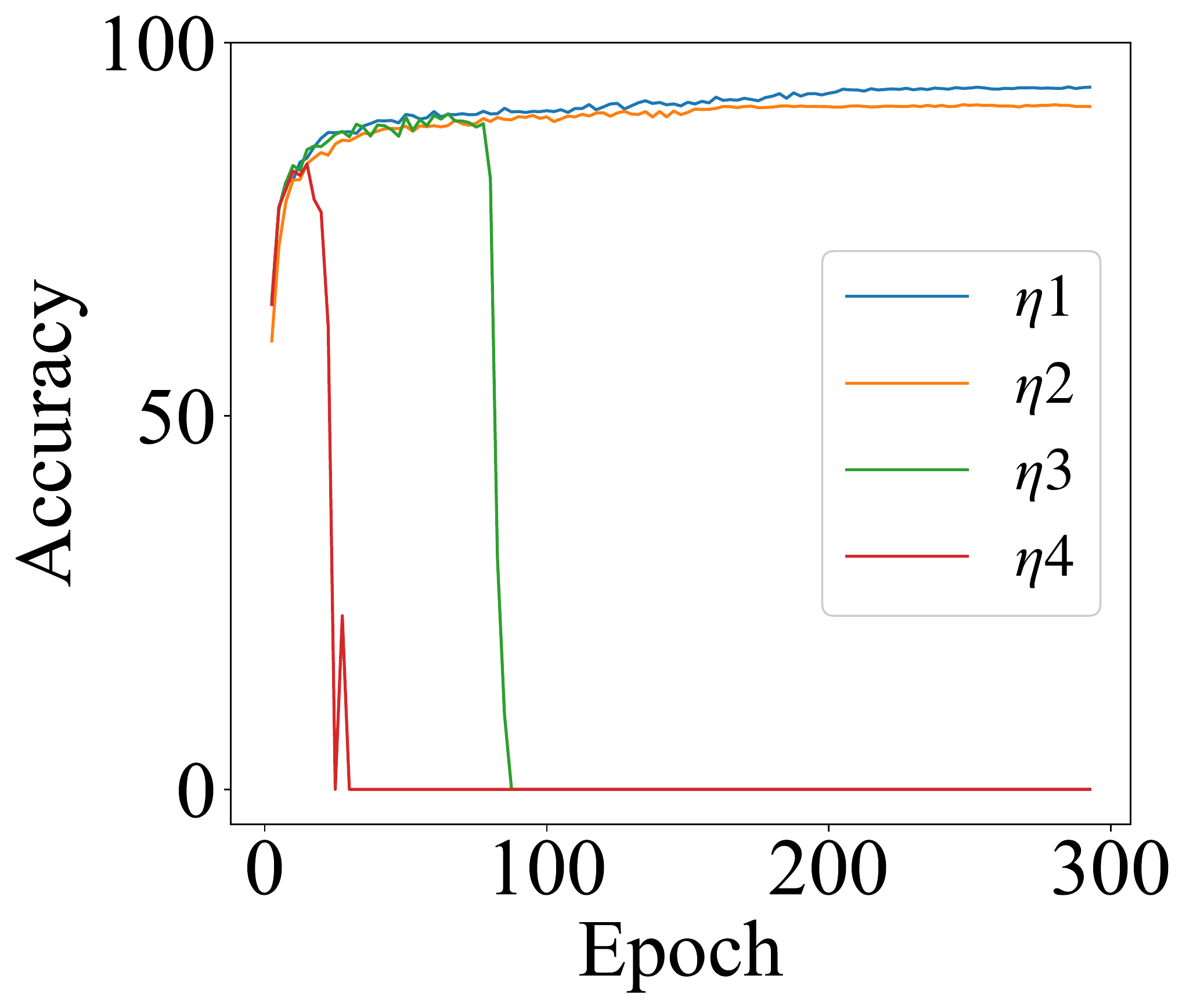}
}
\caption{The effect of clipping and learning rates on INT8 training. $\gamma$  in (a) represents optimal clipping value. In (b), $\eta1$ sets initial learning rate as 0.1 with $\phi(d_c)$ scaling, $\eta2$, $\eta3$ and $\eta4$ choose 0.01, 0.05, 0.1 as initial learning rate respectively without scale.}\label{fig:effect}
\end{figure}

The theoretical analysis on convergence ability of quantized training indicates the necessity of scaling down learning rate, since the quantization error of gradients cannot vanish completely. To validate this point, we decrease the learning rate of the original crashed training of MobileNetV2 mentioned in Section \ref{section:challenges} and find that it defers and even eliminates the crash with an extremely low learning rate, although facing a performance degradation (see the red, green and orange lines in Figure \ref{fig:effect}(b)). 

Since the gradients are backward propagated layer by layer, the minor gradient deviation will accumulate exponentially after massive multiplication and addition calculation. To address this issue, we further propose the Deviation Counteractive Learning Rate Scaling to balance out the error by exponentially decaying the learning rate according to the degree of direction deviation $d_c$, the scaling function is formulated at:
\begin{equation}
    \phi(d_c) = \max(e ^ {-\alpha d_c}, \beta)
\end{equation}
where $\alpha$ controls the decay degree and $\beta$ limits the lower bound of scaling. 

This scaling function generates a factor to scale down the original full-precision learning rate. We empirically find that the self-adapting scaling function performs well in a layer-wise way, adaptively adjusting the learning rate according to the direction deviations in different layers. This counteracts the undesired effects of the gradient deviations across layers, and exactly addresses the challenges of the depth-specific and structure-specific patterns as observed in characteristics C3 and C4 in Section \ref{section:challenges}. The blue line in Figure \ref{fig:effect}(b) demonstrates that the training equipped with $\phi(d_c)$ scaling achieves higher accuracy than the manually adjusted ones (tested with MobileNetV2 on CIFAR-10).


\begin{figure}[t!]
\centering
\includegraphics[width=1\linewidth]{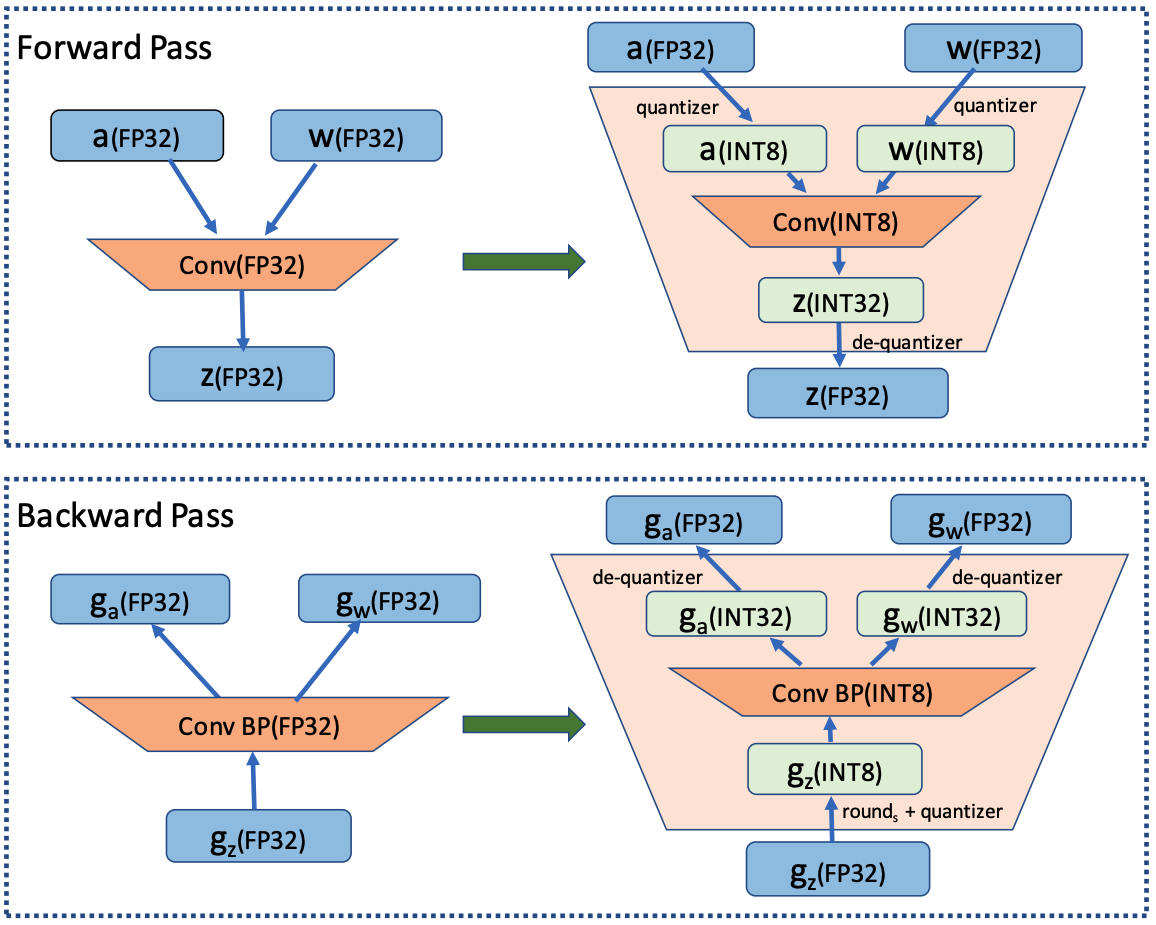}
\caption{Flexible INT8 convolutional layer replacement.}
\label{fig:replace}
\end{figure}

\subsection{General Purpose Training Framework}
\begin{table}[t!]
\caption{Overhead reduced with Periodic Update (on ResNet-50).}
\label{table:period}
\centering
\small
\begin{tabular}{|c|c|c|c|c|}
\hline
 Period & 1  & 10  & 100 & 1000 \\
\hline
\hline
 Average time(s/iter)& 1.006 & 0.364 & 0.301 & 0.297 \\
\hline
\end{tabular}
\vspace{-0.1in}
\end{table}
In addition to ensuring the stable and accurate convergence, in practice our unified INT8 training framework should also satisfy the following three features:

\noindent (1) \textbf{Easy to plug into any DCNN architecture.} To realize this, we implement an automatic match and replacement mechanism in PyTorch \cite{pytorch} that correspondingly substitutes convolutional and fully-connected layers with 8-bit counterpart. The whole workflow including both forward and backward passes is shown in Figure \ref{fig:replace}.

\noindent (2) \textbf{No excessive extra computational overhead.} To avoid the extra time cost of calculating clipping value, we design a Periodic Update method to optimize the clipping value periodically. As we can see in Table \ref{table:period}, the Periodic Update method dramatically reduces the computational overhead of optimizing the clipping value. 

\noindent (3) \textbf{Easy to implement on off-the-shelf hardware.} To validate the potential of that, we utilizes the DP4A instruction (8-bit integer 4-element vector dot product) on low-end NVIDIA Pascal GPUs to implement efficient 8-bit kernels for calculating gradients. To the best of our knowledge, we are the first to achieve practical acceleration of INT8 training including the backward propagation. The detailed speedup will be reported and discussed in Section \ref{section:speed}.

\section{Experiments}
We conduct extensive experiments to demonstrate that our proposed framework is unified for various network structures on popular image classification and object detection tasks with state-of-the-art accuracy, and meanwhile it can be easily deployed on the mainstream devices (NVIDIA Pascal GPU) with satisfactory speedup, compared to full-precision training.

\subsection{Ablation Study}
\noindent \textbf{Settings.} We first conduct the ablation study on CIFAR-10 dataset with MobileNetV2 \cite{MobileNet}, to validate the effectiveness of the proposed techniques. We use cosine scheduler \cite{QSGD} with initial learning rate set to 0.1 for all experiments. In the Periodic Update experiment, the $\alpha$ and $\beta$ in learning rate scaling are set to 20 and 0.1 respectively. 

\begin{figure}[t!]
\centering
\subfigure[the cosine distance]{
\includegraphics[width=.47\linewidth]{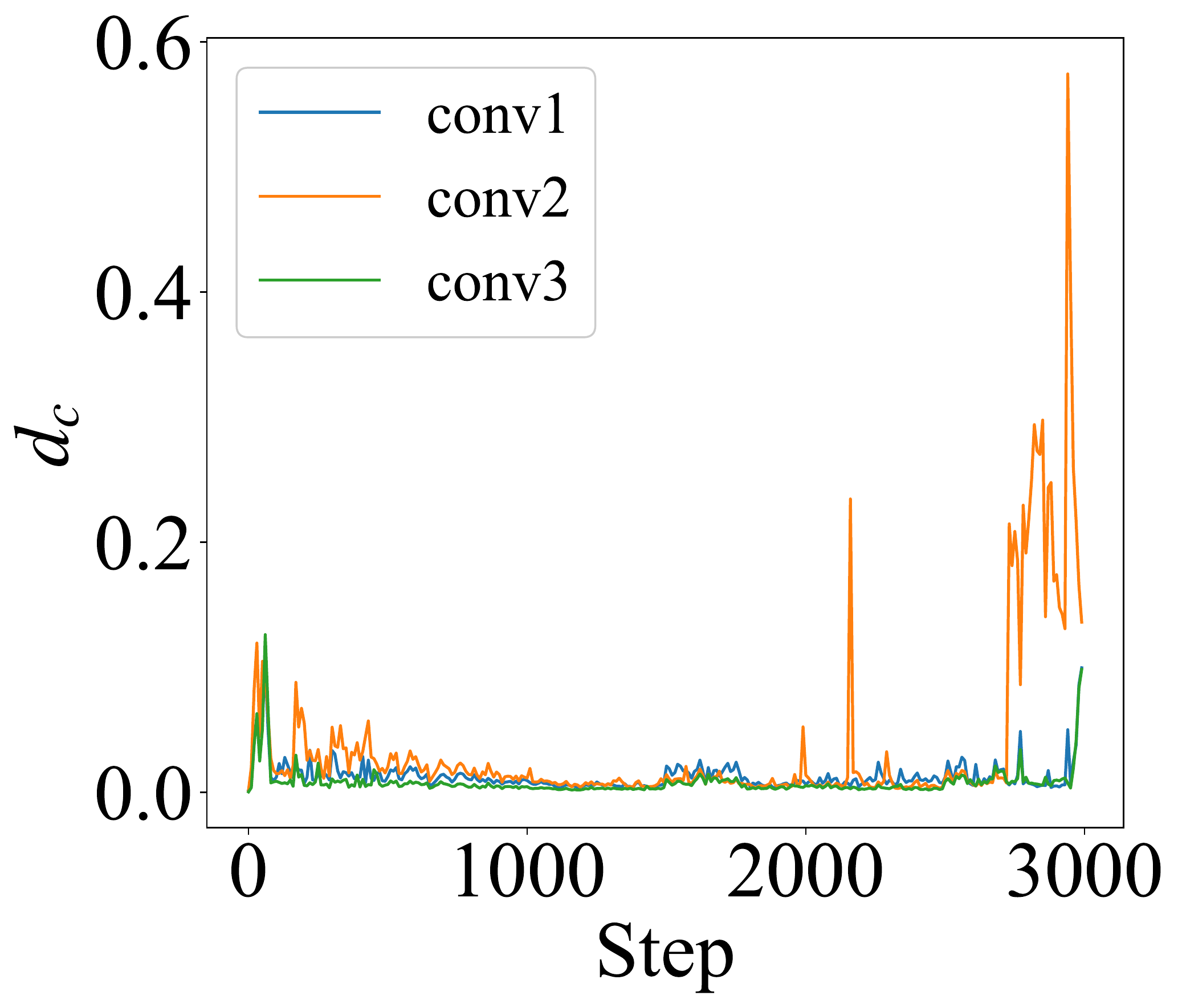}
}
\subfigure[the accuracy curve]{
\includegraphics[width=.47\linewidth]{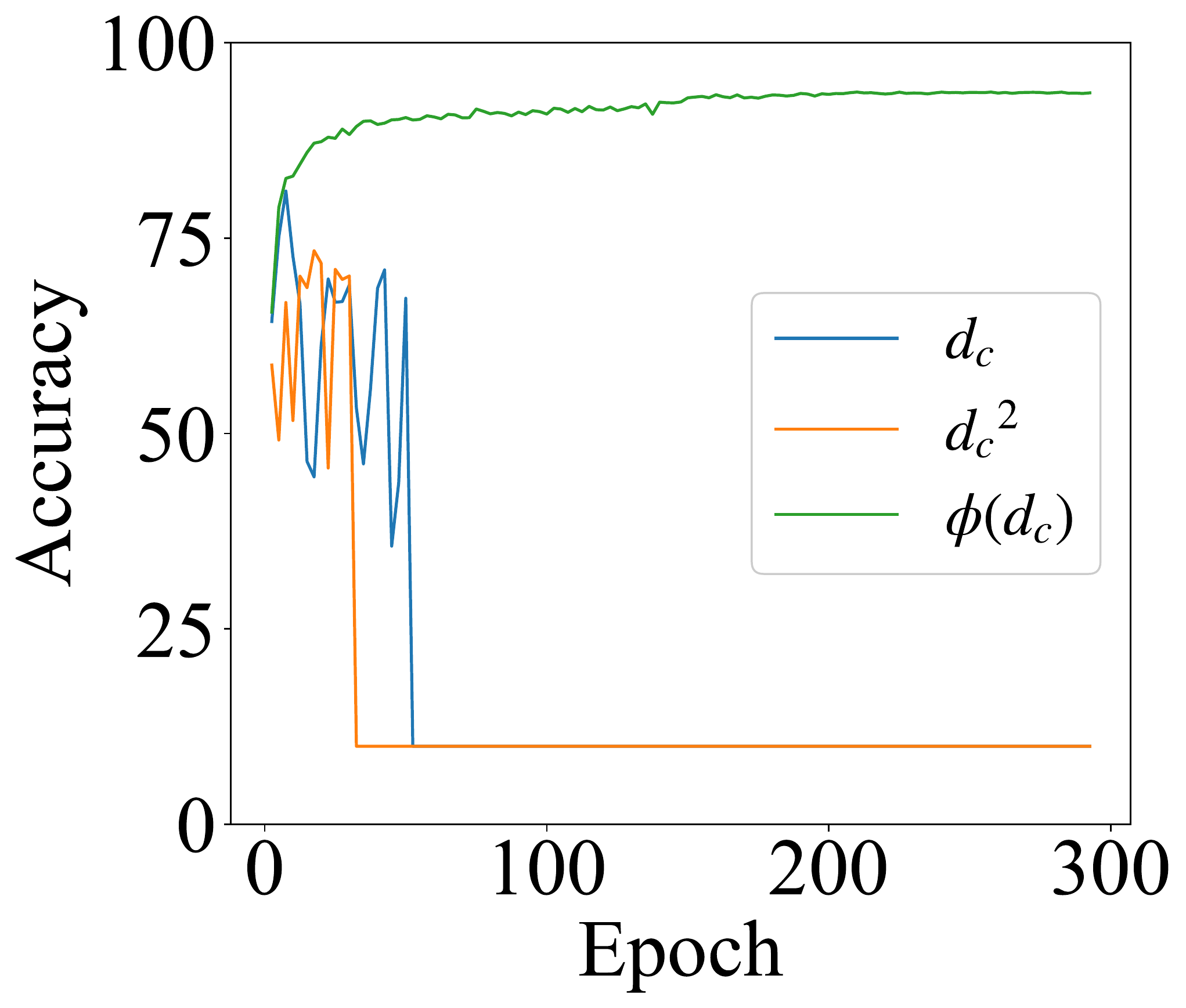}
}
\caption{Analysis of cosine distance and learning rate scaling function.}
\vspace{-0.1in}
\label{fig:cosinedistance}
\end{figure}

\noindent \textbf{Direction Sensitive Gradient Clipping.} Figure \ref{fig:cosinedistance}(a) shows the cosine distance with respect to the training steps. We can observe that conv2 (the second convolutional layer) of each block owns a much larger cosine distance than other layers of the block most of the time. This is consistent with C4 that the gradients of conv2 own sharper shape, indicating that our cosine distance can well reflect the gradient characteristics.

Moreover, as Table \ref{table:clip} lists, our proposed direction sensitive gradient clipping technique indeed prevents INT8 training from crashing, which proves the fact that optimizing a clipping value of gradients to minimize direction deviation $d_c$ can certainly ensure a stable INT8 training.

\begin{table}[htbp!]
\caption{Ablation study on clipping method for INT8 training.}
\label{table:clip}
\centering
\small
\begin{tabular}{|c|c|c|}
\hline
Clipping method & No clipping & \tabincell{c}{Direction Sensitive \\Gradient Clipping} \\
\hline
\hline
    Accuracy (\%) &  NaN & 93.02 \\
\hline
\end{tabular}
\vspace{-0.05in}
\end{table}

\noindent \textbf{Deviation Counteractive Learning Rate Scaling.} We evaluate three forms of learning rate scaling strategies without clipping to control variable for a reasonable comparison. The results shown in Figure \ref{fig:cosinedistance}(b) reveal that linear and quadratic forms are too weak to control optimization direction within the convergence boundary and model crashes in the training process. Compared with linear and quadratic form, the scaling with exponential form is more powerful to counteract the direction deviation and prevents optimization from stepping out of the convergence boundary. We further explore its sensitivity to the selection of hyperparameter in Table \ref{table:LRParam}, and we can see that different settings of $\alpha$ and $\beta$ achieve similar accuracy, which presents the stability of our Deviation Counteractive Learning Rate Scaling.

\begin{table}[htbp!]
\vspace{-0.05in}
\centering
\caption{Comparison of different hyperparameters for learning rate scaling.}
\label{table:LRParam}
\centering
\small
\begin{tabular}{|c|c|c|c|c|}
\hline
$\alpha$ & 10 & 10 & 20 & 20 \\
\hline
$\beta$ & 0.1 & 0.2 & 0.1 & 0.2 \\
\hline
\hline
 Accuracy (\%) & 92.82 & 93.28 & 93.38 & 93.27 \\
\hline
\end{tabular}
\vspace{-0.05in}
\end{table}


\noindent \textbf{Periodic Update for clipping value.} To reduce the extra computational overhead, we increase the period to update clipping value and find that it brings little hurt to the accuracy, as shown in Table \ref{table:Periodic}. This empirical conclusion brings possibilities for the practical acceleration of INT8 training. Besides, here we apply both gradient clipping and learning rate scaling, and obtain better performance (see that with period 1) than those in Table \ref{table:clip} and \ref{table:LRParam}. This further verifies the positive effects of the two general techniques.

\begin{table}[htbp!]
\vspace{-0.05in}
\caption{Ablation study on update period.}
\label{table:Periodic}
\centering
\small
\begin{tabular}{|c|c|c|c|c|}
\hline
 Period & 1 & 10 &  100 & 1000  \\
\hline
\hline
  Accuracy (\%) &  93.66 & 93.07 & 93.38 & 92.75 \\
\hline
\end{tabular}
\end{table}

\subsection{Image Classification}
Now we consider the popular image classification task that most prior studies choose to evaluate the quantization performance. We experiment with AlexNet \cite{AlexNet}, ResNet \cite{Resnet}, MobileNetV2 \cite{MobileNet} and InceptionV3 \cite{Inception} on CIFAR-10 \cite{CIFAR} and ImageNet (ILSVRC2012) \cite{Imagenet}. The CIFAR-10 dataset contains a training set of 50K images and a testing set of 10k images. Each image is of size 3$\times$3 with 10 classes. ImageNet (ILSVRC2012) consists of 1.2 million training images and 50K test images with 1000 classes. 

\noindent \textbf{Settings.} As for the hyperparameters of ResNet, we use the same settings described in \cite{Resnet}. For other neural networks, we use cosine scheduler \cite{QSGD} with initial learning rate set to 0.1. The $\alpha$ and $\beta$ in learning rate scaling are set to 20 and 0.1 respectively. Clipping value is updated per 100 iterations for all experiments.    

\noindent \textbf{CIFAR-10.} As Table~\ref{table:CIFAR-10} shows, our method achieves comparable accuracy on ResNet-20 to FP8 training, but takes much less memory and computation consumption due to the fixed-point operation. Moreover, our method performs surprisingly good on MobileNetV2 (1.01$\%$ accuracy drop) and InceptionV3 (even better than full precision model). 

\noindent \textbf{ImageNet.} Table~\ref{table:IMAGENET} lists existing state-of-the-art quantized training methods including WAGE \cite{wage}, WAGEUBN \cite{wageubn} and FP8 training \cite{fp8training}. For AlexNet INT8 training, our method obtains 5.84\% improvement over DoReFa-Net \cite{zhou2016dorefa}. Free from the extra overhead like $\tanh$, our method enjoys higher efficiency than DoReFa-Net. As for the 2-bit weight and 8-bit activation/gradient case, we significantly outperform WAGE with about 3\% accuracy gain. What's more, equipped with our method, the INT8 training for ResNet architecture achieves almost no performance degradation, while none of the previous studies has done that. Compared with the FP8 training method, our method improves the accuracy by nearly 3\%. It should be noted that we can directly get a real speedup on popular off-the-shelf devices while methods like FP8 training need specially designed hardware, which means that our framework is more general for unified training acceleration.

As analyzed in \cite{li2016performance}, the convolutional layer occupies most of the training time while other layers like BatchNorm and ReLU are not computation-intensive. Therefore, we mainly focus on quantizing convolutional layers currently and do not quantize BatchNorm layer like RangeBN \cite{banner2018scalable} and WAGEUBN \cite{wageubn}. Even so, there is still a significant speedup for INT8 training. In addition, we could get comparable accuracy to full precision training, much higher than RangeBN and WAGEUBN.

\noindent \textbf{Networks using INT8 training for the first time.} To our best knowledge, we are the first to quantize gradient of MobileNetV2, which is known to be difficult in this community. Our method gets very good performance on both CIFAR-10 and ImageNet datasets using MobileNetV2, with only around 1$\%$ accuracy loss. We also try INT8 training on InceptionV3 for the first time, and achieve comparable accuracy to full precision model. Note that for InveptionV3 on CIFAR-10, our INT8 training method can even achieve better performance than the full-precision model.

\begin{table}[t!]
\caption{Results on CIFAR-10 dataset.}
\label{table:CIFAR-10}
\centering
\small
\begin{tabular}{|c|c|c|c|}
\hline
Model & Method & \tabincell{c}{Bit-width \\ (W/A/G)} & \tabincell{c}{Accuracy \\ (\%)} \\
\hline
\hline
\multirow{3}{*}{ResNet-20}
            & FP & 32/32/32 & 92.32 \\
            \cline{2-4}
            & FP8 training \cite{fp8training} & 8/8/8 & 92.21 \\
            \cline{2-4}
            & Ours & 8/8/8 & 91.95 \\
        \hline
        \multirow{2}{*}{\tabincell{c}{MobileNetV2}}
            & FP & 32/32/32 & 94.39 \\
            \cline{2-4}
            & Ours & 8/8/8 & \textbf{93.38} \\
        \hline
        \multirow{2}{*}{\tabincell{c}{InceptionV3}}
            & FP & 32/32/32 & 94.89 \\
            \cline{2-4}
            & Ours & 8/8/8 & \textbf{95.00} \\
\hline
\end{tabular}
\vspace{-0.15in}
\end{table}

\begin{table}[t!]
\caption{Results on ImageNet dataset.}
\label{table:IMAGENET}
\centering
\small
\begin{tabular}{|c|c|c|c|}
\hline
Model & Method & \tabincell{c}{Bit-width \\ (W/A/G)} & \tabincell{c}{Accuracy \\(\%)} \\
\hline
\hline
\multirow{6}{*}{AlexNet}
            & FP & 32/32/32 & 59.84 \\
            \cline{2-4}
            & DoReFa-Net \cite{zhou2016dorefa} & 8/8/8 & 53.00 \\
            & Ours & 8/8/8 & \textbf{58.84} \\
            \cline{2-4}
            & WAGE \cite{wage} & 2/8/8 & 48.40 \\
            & Ours & 2/8/8 & \textbf{51.28} \\
        \hline
        \multirow{4}{*}{ResNet-18}
            & FP & 32/32/32 & 70.30 \\
            \cline{2-4}
            & WAGEUBN \cite{wageubn}  & 8/8/8 & 66.92 \\
            \cline{2-4}
            & FP8 training \cite{fp8training} & 8/8/8 & 67.34 \\
            \cline{2-4}
            & Ours & 8/8/8 & \textbf{69.67} \\
        \hline
        \multirow{3}{*}{ResNet-34}
            & FP & 32/32/32 & 73.68 \\
            \cline{2-4}
            & WAGEUBN \cite{wageubn}  & 8/8/8 & 68.50 \\
            \cline{2-4}
            & Ours & 8/8/8 & \textbf{73.29} \\
        \hline
        \multirow{3}{*}{ResNet-50}
            & FP & 32/32/32 & 76.60 \\
            \cline{2-4}
            & WAGEUBN \cite{wageubn} & 8/8/8 & 69.07 \\
            \cline{2-4}
            & Ours & 8/8/8 & \textbf{76.34} \\
        \hline
        \multirow{2}{*}{\tabincell{c}{MobileNetV2}}
            & FP & 32/32/32 & 72.39 \\
            \cline{2-4}
            & Ours  & 8/8/8 & \textbf{71.20} \\
        \hline
        \multirow{2}{*}{\tabincell{c}{InceptionV3}}
            & FP & 32/32/32 & 72.39 \\
            \cline{2-4}
            & Ours  & 8/8/8 & \textbf{71.20} \\
\hline
\end{tabular}
\vspace{-0.1in}
\end{table}

\subsection{Object Detection}
To prove the versatility of our method, we further conduct experiments with the popular object detection networks including Faster-RCNN \cite{ren2015faster}, RFCN \cite{dai2016rfcn} and RetinaNet \cite{RetinaNet} on two widely used datasets: PASCAL VOC \cite{Pascalvoc} and COCO~\cite{coco}. The PASCAL VOC dataset consists of 11k images with 20 classes. The COCO dataset contains more than 20k images and 80 object categories. Note that we are the first to successfully achieve INT8 training on the object detection task.

\noindent \textbf{Settings.} As for the hyperparameters, we follow the same rules described in \cite{Li_2019_CVPR}. And $\alpha$ and $\beta$ for learning rate scaling are the same as those used in image classification task.

\noindent \textbf{PASCAL VOC.} We test RFCN and Faster R-CNN with different backbones, and find that quantized training equipped with our method only suffers a very slight detection accuracy (mAP) drop. The result of RFCN shows that even for a deeper backbone such as ResNet-101, our INT8 training still maintains almost the same accuracy as full-precision.

\noindent \textbf{COCO.} On the large scale COCO dataset, we experiment with RetinaNet (one-stage) and Faster R-CNN (two-stage). Our method performs stably with less than 1.8$\%$ accuracy degradation on both networks. We find that RetinaNet incurs higher mAP loss than Faster R-CNN, which is inconsistent with the conclusions in the previous study \cite{Li_2019_CVPR}. This may be caused by the fact that the focal loss used in one stage detector is more sensitive to gradient quantization.


\begin{table}[t!]
\caption{Results on PASCAL VOC Dataset.}
\label{table:PASCALVOC}
\centering
\small
\begin{tabular}{|c|c|c|c|c|}
\hline
Model & Backbone & Method & \tabincell{c}{Bit-width \\ (W/A/G)} & mAP (\%) \\
\hline
\hline
\multirow{2}{*}{\tabincell{c}{Faster\\ R-CNN}}
            & ResNet-50 & FP & 32/32/32 & 82.0 \\
            \cline{2-4}
            & ResNet-50 & Ours & 8/8/8 & \textbf{81.9} \\
            \hline
        \multirow{2}{*}{RFCN}
            & ResNet-101 & FP & 32/32/32 & 80.8 \\
            \cline{2-4}
            & ResNet-101 & Ours & 8/8/8 & \textbf{79.1} \\
\hline
\end{tabular}
\vspace{-0.15in}
\end{table}

\begin{table}[t!]
\caption{Results on COCO Dataset.}
\label{table:COCO}
\centering
\small
\begin{tabular}{|c|c|c|c|c|}
\hline
Model & Backbone & Method & \tabincell{c}{Bit-width \\ (W/A/G)} & mAP (\%) \\
\hline
\hline
\multirow{2}{*}{\tabincell{c}{Faster \\ R-CNN}}
            & ResNet-50 & FP & 32/32/32 & 36.2 \\
            \cline{2-4}
            & ResNet-50 & Ours & 8/8/8 & \textbf{34.95} \\
        \hline
        \multirow{2}{*}{RetinaNet}
            & ResNet-50 & FP & 32/32/32 & 36.9 \\
            \cline{2-4}
            & ResNet-50 & Ours & 8/8/8 & \textbf{35.1} \\
\hline
\end{tabular}
\vspace{-0.1in}
\end{table}

\begin{figure}
\includegraphics[width=1.0\linewidth]{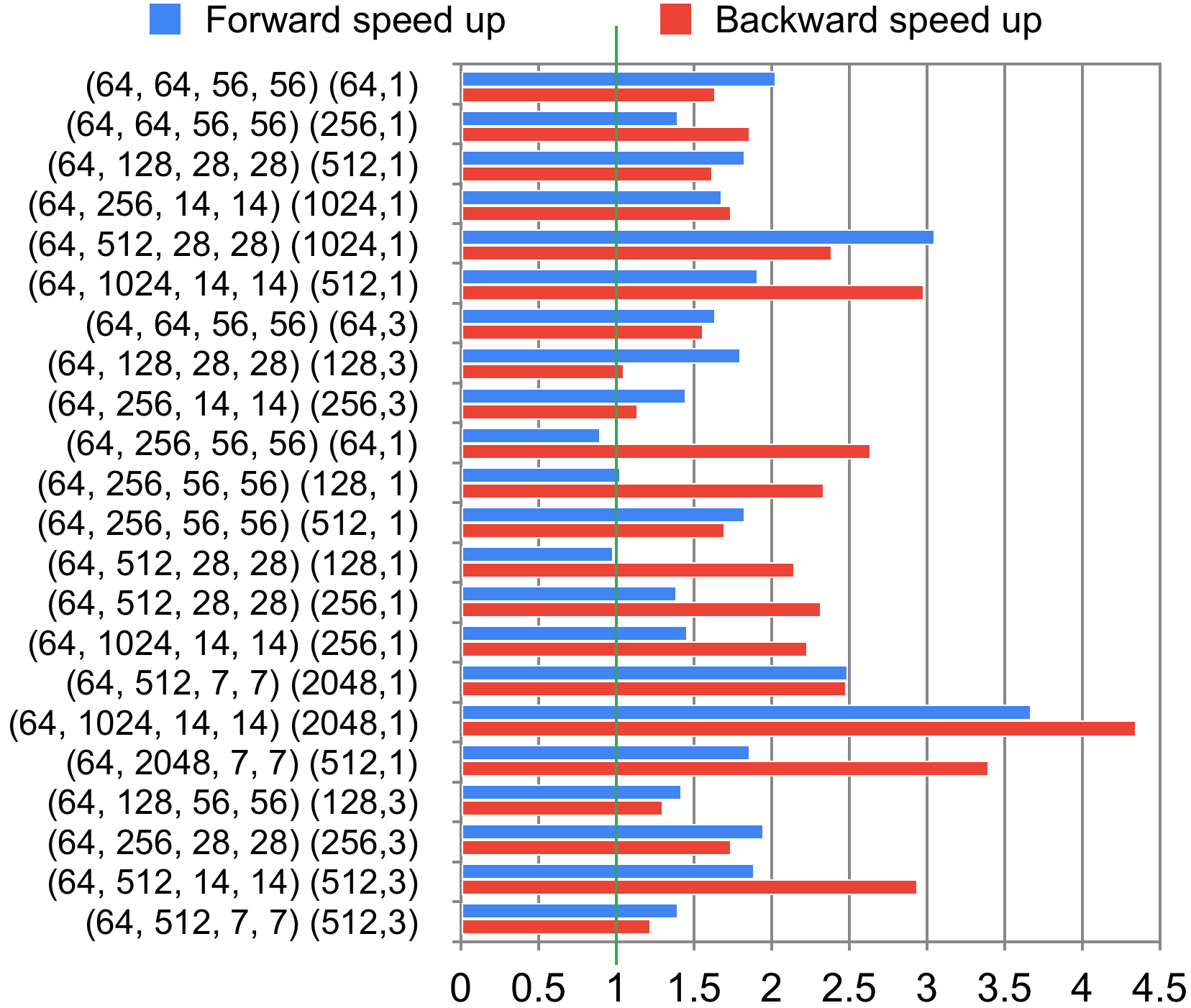}
\caption{INT8 convolution speedup on GPU, where Y-axis indicates (input shape), (kernel number, kernel size) of convolution.}
\label{fig:speedup}
\vspace{-0.05in}
\end{figure}

\begin{table}[t!]
\caption{End-to-end average time for a round of INT8 training. (tested with ResNet-50 on GeForce GTX1080TI, batch size 64.)}
\label{tab:speedup}
\centering
\small
\begin{tabular}{|c|c|c|c|}
\hline
Precision        &  Forward (s) & Backward (s) & Iteration (s) \\ \hline \hline
FP32 (cuDNN)         &   0.117      &   0.221      &    0.360    \\ \hline
INT8 (ours)        &   0.101      &   0.171      &    0.293    \\ \hline

\end{tabular}
\vspace{-0.2in}
\end{table}

\subsection{Speed Result on NVIDIA GPU}
\label{section:speed}
None of the existing libraries can directly support the complete INT8 training. Thus we implement it by ourselves on NVIDIA Pascal GPU using DP4A instruction to verify the acceleration power of our method. The speedup of each convolutional layer in ResNet-50 is shown in Figure \ref{fig:speedup}. In the forward process using our solution, INT8 can bring an average \textbf{1.63$\times$} speedup, while in the backward process, it can achieve a higher \textbf{1.94$\times$} speedup.
Table \ref{tab:speedup} further reports the time consumption and speed improvement of each training round. Even if we only replace the FP32 convolutional layer with the slightly optimized INT8 one, the training time for ResNet-50 can be reduced by about 22\%.

\section{Conclusions}
In this paper, we attempt to build an INT8 training framework for common DCNNs. We found four distinctive characteristics of gradients and then gave two theoretical principles stabilizing training with the convergence bound. Based on that, we proposed Direction Sensitive Gradient Clipping and Deviation Counteractive Learning Rate Scaling. Extensive experiments prove the versatility of our method for various networks and tasks. We reduced the training time by 22\% on Pascal GPU with only trivial optimization. If each layer is sufficiently optimized, the training will achieve higher speedup and lower memory consumption. We hope our first successful attempt can help lead the community towards a fully unified INT8 training.
\nocite{caffe}

{\small
\bibliographystyle{ieee_fullname}
\bibliography{egbib}
}

\clearpage

\section{Supplementary Material}

\subsection{Proof of Theorem 1}

\setcounter{assumption}{0}
\begin{assumption}  \label{assumption:3} 
	$f_t$ is convex;
\end{assumption}
\begin{assumption}  \label{assumption:4} 
	$\forall \mathbf{w}_p, \mathbf{w}_q \in \mathbb{S}, \|\mathbf{w}_p-\mathbf{w}_q \|_\infty \leq D_\infty$.
\end{assumption}

\begin{proof}
	Considering the update for $i$th entry of weight,
	\begin{equation}
	w_{t+1,i} = w_{t,i} - \eta_{t,i}\hat{g}_{t,i}
	\end{equation}
	we have 
	\begin{equation}
	\begin{split}
	&(w_{t+1,i}-w_i^*)^2 = (w_{t,i} - \eta_{t,i}\ \hat{g}_{t,i}-w_i^*)^2 \\
	&\quad =(w_{t,i}-w_i^*)^2 - 2(w_{t,i}-w_i^*)\eta_{t,i}\ \hat{g}_{t,i}+\eta_{t,i}^2\ \hat{g}_{t,i}^2
	\end{split}
	\end{equation}
	Rearrange the equation, and divide $2\eta_{t,i}$ on both side as $\eta_{t,i}$ is none-zero,
	\begin{equation}
	\label{eq:rearrange}
	\begin{split}
	\hat{g}_{t,i}(w_{t,i}-w_i^*) &= \frac{1}{2\eta_{t,i}}(w_{t,i}-w_i^*)^2+\frac{\eta_{t,i}}{2}\hat{g}_{t,i}^2 \\
	&\quad-\frac{1}{2\eta_{t,i}}(w_{t+1,i}-w_i^*)^2
	\end{split}
	\end{equation}
	The error of quantized gradients is defined as
	\begin{equation}
	\epsilon_{t,i} = g_{t,i}-\hat{g}_{t,i}
	\end{equation}
	Replace $\hat{g}_{t,i}$ in the \eqref{eq:rearrange} with $g_{t,i}$ and $\epsilon_{t,i}$, and we can get that
	\begin{equation}
	\label{eq:substitude}
	\begin{split}
	&g_{t,i}(w_{t,i}-w_i^*) = \\
	&\quad \frac{1}{2\eta_{t,i}}[(w_{t,i}-w_i^*)^2-(w_{t+1,i}-w_i^*)^2)]\\
	&\quad + \epsilon_{t,i}(w_{t,i}-w_i^*) + \frac{\eta_{t,i}}{2}(g_{t,i}-\epsilon_{t,i})^2
	\end{split}
	\end{equation}
	According to assumption \ref{assumption:1},
	\begin{equation}
	\label{eq:convex_conclusion}
	f_t(\mathbf{w}_t) - f_t(\mathbf{w}^*) \leq \mathbf{g}_t^\top(\mathbf{w}_t-\mathbf{w}^*)
	\end{equation}
	So combine the \eqref{eq:substitude} and \eqref{eq:convex_conclusion}, sum over the $d$ dimensions of $\mathbf{w}$ and the $T$ iterations, then the regret
	\begin{equation}
	\label{ieq:regret}
	\begin{split}
	R(T) &\leq \sum_{t=1}^{T} \sum_{i=1}^{d} (\frac{1}{2\eta_{t,i}} [(w_{t,i}-w_i^*)^2 - (w_{t+1,i}-w_i^*)^2] \\
	&\quad \quad \quad \quad \quad + \epsilon_{t,i}(w_{t,i}-w_i^*) + \frac{\eta_{t,i}}{2}(g_{t,i}-\epsilon_{t,i})^2) \\
	&= \sum_{i=1}^{d} [\frac{1}{2\eta_{1,i}}(w_{1,i}-w_i^*)^2 - \frac{1}{2\eta_{T,i}}(w_{T+1,i}-w_i^*)^2] \\
	&\quad + \sum_{t=2}^{T}\sum_{i=1}^{d}(\frac{1}{2\eta_{t,i}}-\frac{1}{2\eta_{t-1,i}})(w_{t,i}-w_i^*)^2 \\
	&\quad + \sum_{t=1}^{T} \sum_{i=1}^{d} [\epsilon_{t,i}(w_{t,i}-w_i^*) + \frac{\eta_{t,i}}{2}(g_{t,i}-\epsilon_{t,i})^2]
	\end{split}
	\end{equation}
	Combine \eqref{ieq:regret} with the assumption \ref{assumption:2}, and we can further relax the above \eqref{ieq:regret} to
	\begin{equation}
	\begin{split}
	R(T) &\leq \sum_{i=1}^{d} \frac{D_\infty^2}{2\eta_{1,i}} + \sum_{t=2}^{T}\sum_{i=1}^{d}(\frac{1}{2\eta_{t,i}}-\frac{1}{2\eta_{t-1,i}})D_\infty^2\\
	&\quad + \sum_{t=1}^{T} \sum_{i=1}^{d} [\epsilon_{t,i}(w_{t,i}-w_i^*) + \frac{\eta_{t,i}}{2}(g_{t,i}-\epsilon_{t,i})^2]
	\end{split}
	\end{equation}
	Assume that all layers have the same learning rate, then
	\begin{equation}
	\begin{split}
	R(T) &\leq \frac{d\ D_\infty^2}{2\eta_{T}} + \sum_{t=1}^{T} \mathbf{\epsilon_t}(\mathbf{w_t}-\mathbf{w^*}) + \sum_{t=1}^{T} \frac{\eta_{t}}{2}(\mathbf{g_{t}}-\mathbf{\epsilon_{t}})^2
	\end{split}
	\end{equation}
	Based on Cauchy's inequality and assumption \ref{assumption:2}, we finally get
	\begin{equation}
	\begin{split}
	R(T) &\leq \frac{d\ D_\infty^2}{2\eta_{T}} + \sum_{t=1}^{T} \| \mathbf{\epsilon_t \|}\cdot \| \mathbf{w_t}-\mathbf{w^*} \| + \sum_{t=1}^{T} \frac{\eta_{t}}{2}\|\mathbf{g_{t}}-\mathbf{\epsilon_{t}}\|^2 \\
	&\leq \frac{d\ D_\infty^2}{2\eta_{T}} + D_\infty \sum_{t=1}^{T} \| \mathbf{\epsilon_t \|} + \sum_{t=1}^{T} \frac{\eta_{t}}{2}\|\mathbf{\hat{g}_{t}}\|^2 \\
	\end{split}
	\end{equation}
	Thus the average regret
	\begin{equation}
	\label{eq:regret}
	\frac{R(T)}{T} \leq \underbrace{\frac{d\ D_\infty^2}{2T\eta_{T}}\vphantom{\sum_{t=1}^{T}}}_{(1)} + \underbrace{ \frac{D_\infty}{T} \sum_{t=1}^{T} \| \mathbf{\epsilon_t} \|\vphantom{\sum_{t=1}^{T}}}_{(2)} + \underbrace{\frac{1}{T}\sum_{t=1}^{T} \frac{\eta_{t}}{2}\|\mathbf{\hat{g}_{t}}\|^2}_{(3)}
	\end{equation}
\end{proof}

\subsection{INT8 Training Stability}

\begin{figure}[htbp!]
	\centering
	\subfigure[the accuracy curve]{
		\includegraphics[width=.47\linewidth]{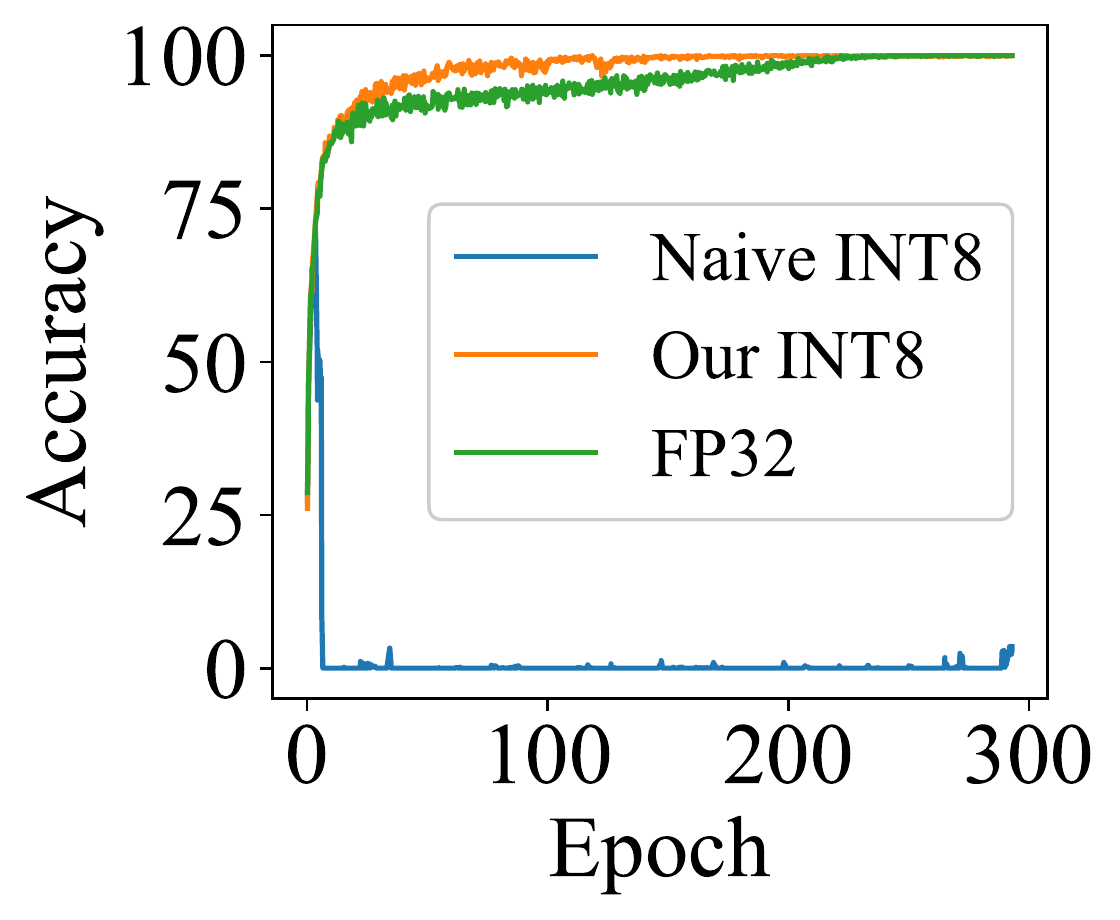}
	}
	\subfigure[the loss curve]{
		\includegraphics[width=.47\linewidth]{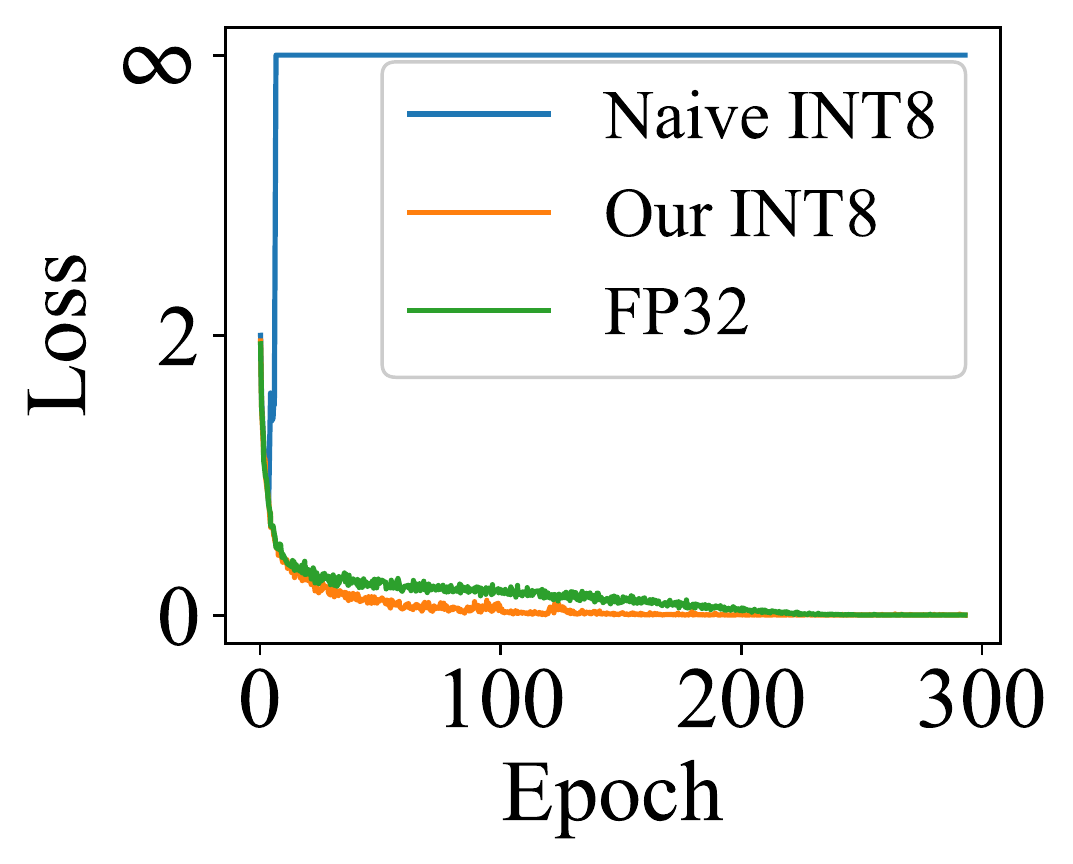}
	}
	\caption{Comparison of INT8 training and FP32 training on CIFAR-10 using MobileNetV2.}
	\label{fig:compare}
	\vspace{-0.1in}
\end{figure}

\begin{figure}[htbp!]
	\centering
	\subfigure[the accuracy curve]{
		\includegraphics[width=.47\linewidth]{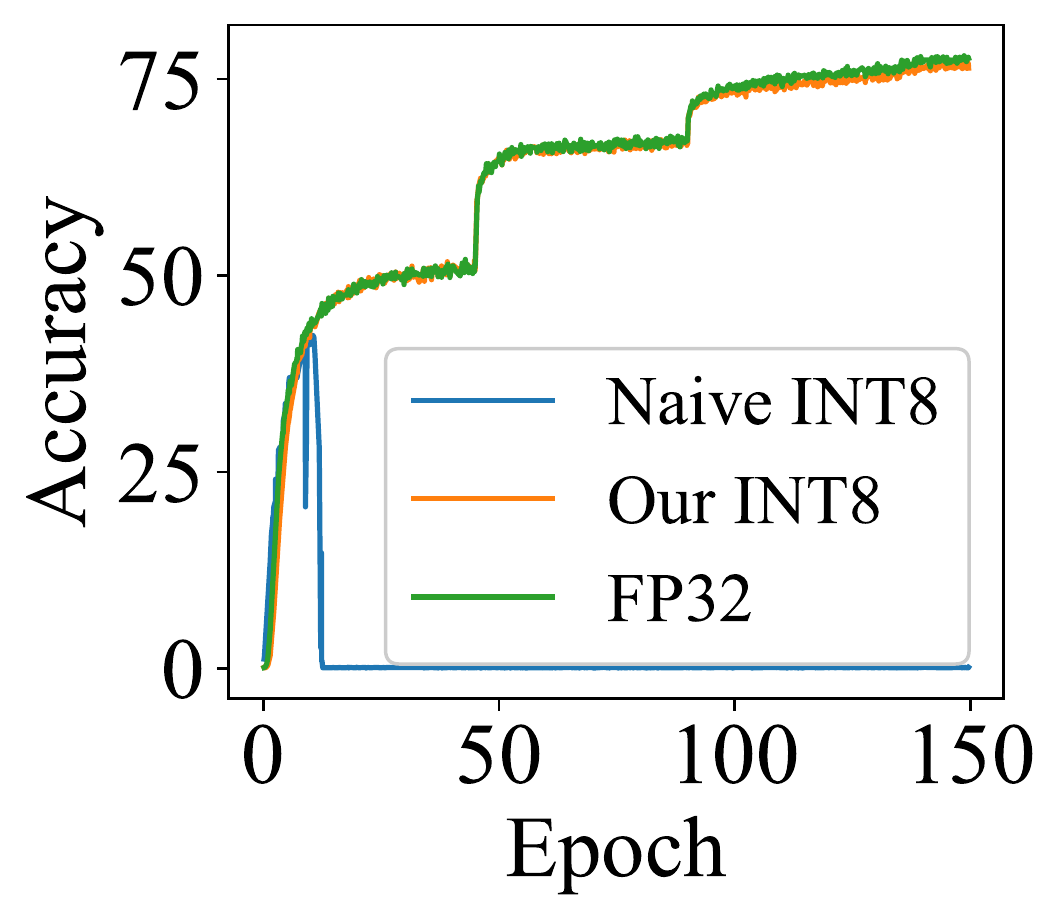}
	}
	\subfigure[the loss curve]{
		\includegraphics[width=.47\linewidth]{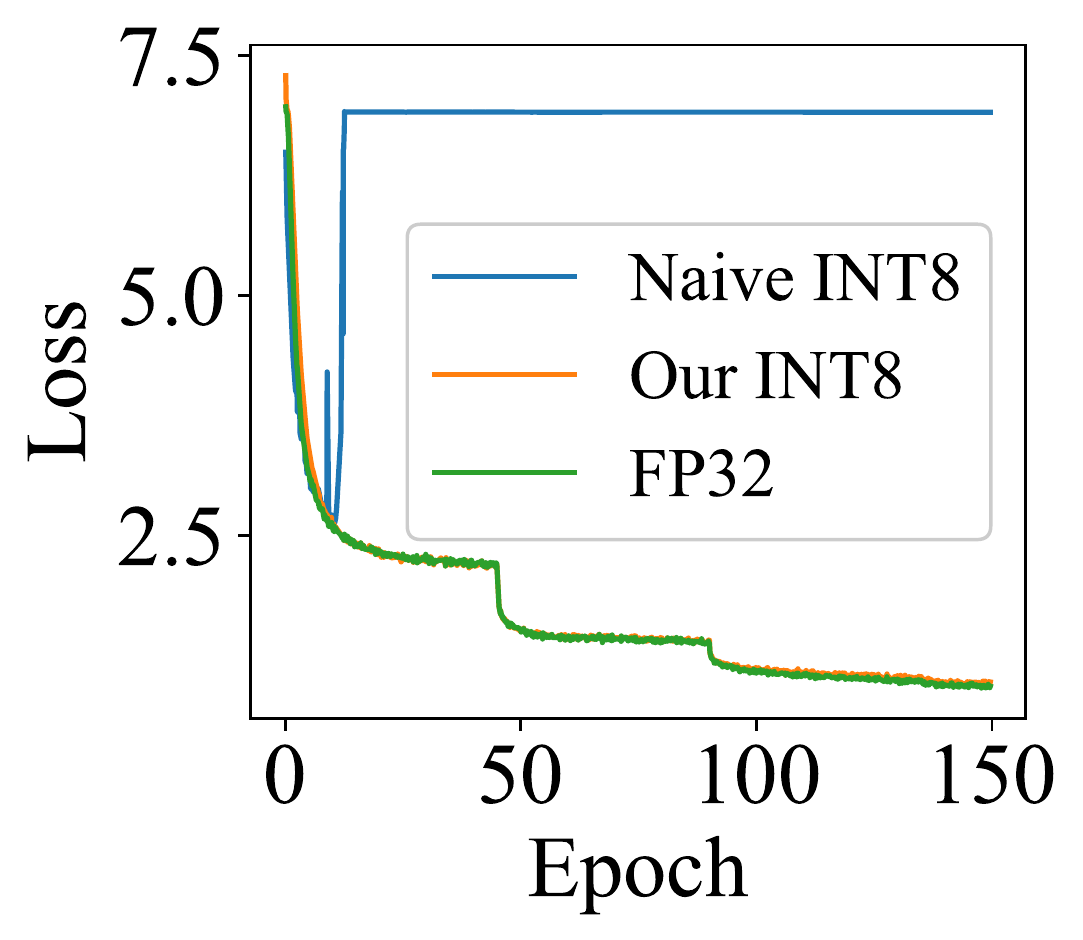}
	}
	\caption{Comparison of INT8 training and FP32 training on ImageNet using ResNet-50.}
	\label{fig:compare_res50}
	\vspace{-0.1in}
\end{figure}

We plot the accuracy and the loss curve of MobileNetV2 training on CIFAR-10 dataset and ResNet-50 training on ImageNet dataset to show the stability of INT8 training. From Figure \ref{fig:compare} and Figure \ref{fig:compare_res50}, we can see that our method makes INT8 training smooth and achieves accuracy comparable to FP32 training. The quantization noise increases exploratory ability of INT8 training since the quantization noise at early stage of training could make the optimization direction more diverse, and with properly reduced learning rate, INT8 training sometimes even converge faster than FP32 training.

\subsection{INT8 Convolution Speed Up Algorithm}
\subsubsection{INT8 Convolution}
On NVIDIA GPUs with Pascal architectures (such as GP102, GP104, and GP106), the new 8-bit integer 4-element dot product with accumulation (DP4A) \cite{ptx} instruction is supported. This enables the NVIDIA GeForce GTX 1080Ti (based on GP102) to achieve a peak integer throughput of 44 Tera Operations Per Second (TOPS), while the peak float throughput is only 11 Tera Float Operations Per Second (TFLOPS).

\begin{figure}[t!]
	\centering
	\includegraphics[width=1\linewidth]{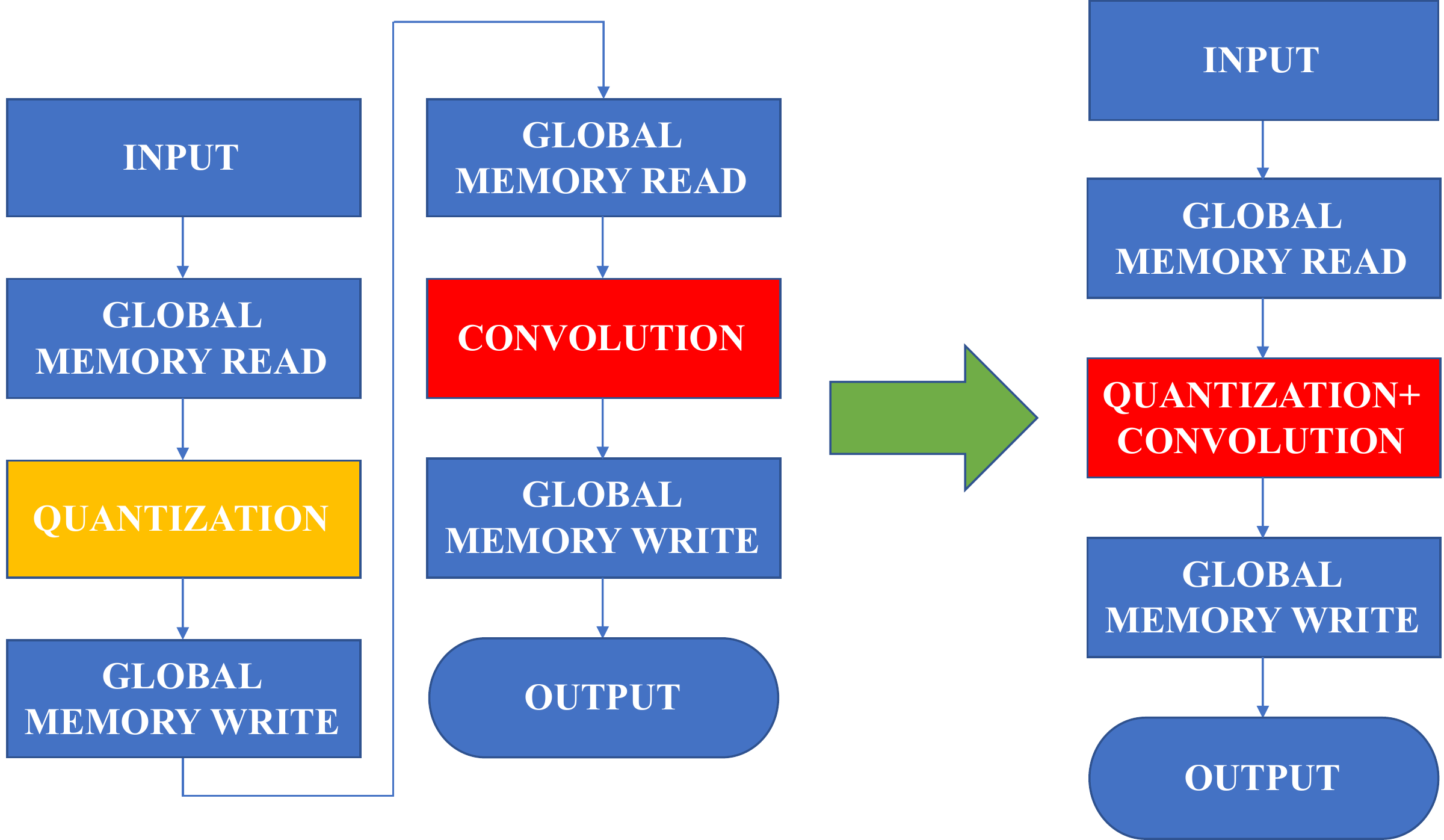}
	\caption{Quantization-convolution fused kernel avoids one extra global memory read and write.}
	\label{fig:quant_conv}
	\vspace{-0.1in}
\end{figure}

Since the release of cuDNN 6.0 \cite{cudnn}, INT8 inference is supported but the INT8 backward process is not implemented. So we use the DP4A instruction to implement the INT8 backward process by ourselves. Moreover, we find that the quantization process before INT8 convolution computation is pretty time-consuming as the quantization needs to read and write the whole data. In order to reduce the overhead that quantization brings, we fuse the quantization process with the convolution computation (quantization-convolution fused kernel). In Figure \ref{fig:quant_conv}, we can see that the combination of quantization and convolution could avoid one extra global memory read and write effectively. Thus we rewrite the INT8 forward and backward process using this quantization-convolution fused kernel and achieve a significant speed-up. 

\begin{figure}[t!]
	\centering
	\includegraphics[width=1\linewidth]{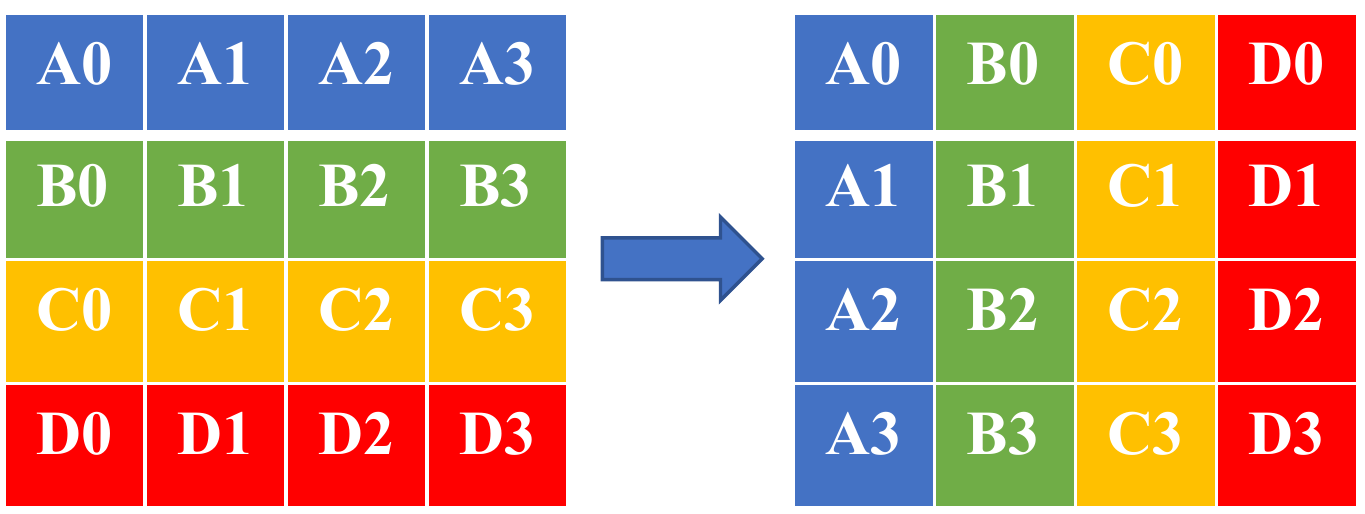}
	\caption{4$\times$4 8-bit integer block transpose in a thread using $prmt$ instruction.}
	\label{fig:transpose}
	\vspace{-0.1in}
\end{figure}

In our implementation, we transpose the data layout into NC4HW so that we can use the DP4A instruction to conduct the convolution computation. We use the $prmt$ instruction in Parallel Thread Execution and Instruction Set Architecture (PTX ISA) \cite{ptx} to transpose the data efficiently. This $prmt$ instruction picks four arbitrary bytes from two 32-bit registers, and reassembles them into a 32-bit destination register. Figure \ref{fig:transpose} shows that one thread can transpose data in 4$\times$4 8-bit integer block by using 12 $prmt$ instructions with shared memory. The transpose implementation code is listed below.
\begin{lstlisting}
int regLDG[4]; int4 regPRMT; int tmp;
asm volatile("prmt.b32 %0, %1, %2, 0x0040;" : "=r"(regPRMT.x) : "r"(regLDG[0]), "r"(regLDG[1]));
asm volatile("prmt.b32 %0, %1, %2, 0x0040;" : "=r"(tmp) : "r"(regLDG[2]), "r"(regLDG[3]));
asm volatile("prmt.b32 %0, %1, %2, 0x5410;" : "=r"(regPRMT.x) : "r"(regPRMT.x), "r"(tmp));
asm volatile("prmt.b32 %0, %1, %2, 0x0051;" : "=r"(regPRMT.y) : "r"(regLDG[0]), "r"(regLDG[1]));
asm volatile("prmt.b32 %0, %1, %2, 0x0051;" : "=r"(tmp) : "r"(regLDG[2]), "r"(regLDG[3]));
asm volatile("prmt.b32 %0, %1, %2, 0x5410;" : "=r"(regPRMT.y) : "r"(regPRMT.y), "r"(tmp));
asm volatile("prmt.b32 %0, %1, %2, 0x0062;" : "=r"(regPRMT.z) : "r"(regLDG[0]), "r"(regLDG[1]));
asm volatile("prmt.b32 %0, %1, %2, 0x0062;" : "=r"(tmp) : "r"(regLDG[2]), "r"(regLDG[3]));
asm volatile("prmt.b32 %0, %1, %2, 0x5410;" : "=r"(regPRMT.z) : "r"(regPRMT.z), "r"(tmp));
asm volatile("prmt.b32 %0, %1, %2, 0x0073;" : "=r"(regPRMT.w) : "r"(regLDG[0]), "r"(regLDG[1]));
asm volatile("prmt.b32 %0, %1, %2, 0x0073;" : "=r"(tmp) : "r"(regLDG[2]), "r"(regLDG[3]));
asm volatile("prmt.b32 %0, %1, %2, 0x5410;" : "=r"(regPRMT.w) : "r"(regPRMT.w), "r"(tmp));
\end{lstlisting}

After transposition, we use two kinds of algorithms im2col plus GEMM \cite{caffe,im2col} and implicit GEMM \cite{cudnn} to implement convolution, and choose a faster algorithm for each convolution layer before training. Through these two algorithms, we convert the original convolution into dot product. Then we use one float load instruction to load four INT8 data and one DP4A instruction to compute four INT8 dot product operations. This can speed up the INT8 convolution significantly.

\subsubsection{Stochastic Rounding}
Due to the use of stochastic rounding in quantizing gradients, we need to generate uniform random numbers during the backward process. One way to generate random numbers is using $curandGenerator$, but this instruction needs extra global memory access, which will significantly degrade our INT8 convolution performance, with time consumption increasing over 100$\%$. Another method is to use $curand\_uniform$, and we need to set a unique $curandState$ for each thread to get different random numbers, which requires a large amount of gpu memory. Worse still, this method runs as slow as the first method. Considering both disadvantages above, we use Linear Congruential Generator (LCG) \cite{lcg} to yield a sequence of pseudo-randomized numbers instead.

The generator is defined by recurrence relation,
\begin{equation}
\label{eq:random}
X_{n+1} = (aX_n + c) \bmod m,
\end{equation}
where $X$ is the sequence of pseudo-random values, $m$ is the modules, $a$ is the multiplier, $c$ is the increment, and $X_0$ is the random seed. The parameters $a$, $c$ and $m$ are set to constants. 

In order to get different random seeds in each thread, we set the random seed $X_0$ to first input data and add the thread index to $X_0$. With above settings, each thread can get a unique random seed. The LCG method generates random numbers quickly and brings slight time consumption to INT8 convolution. 

\end{document}